\documentclass{article}

\usepackage{microtype}
\usepackage{graphicx}
\usepackage{subfigure}
\usepackage{booktabs} 
\usepackage{ragged2e}
\usepackage{multirow}   
\usepackage{xspace}
\usepackage{rotating}
\usepackage{float}
\usepackage{sidecap}
\usepackage[breakable]{tcolorbox}

\usepackage{hyperref}

\usepackage[accepted]{icml2025}

\usepackage{amsmath}
\usepackage{amssymb}
\usepackage{mathtools}
\usepackage{amsthm}
\usepackage{acronym}

\usepackage[capitalize,noabbrev]{cleveref}

\theoremstyle{plain}

\theoremstyle{definition}

\theoremstyle{remark}

\newcommand{\method}{\texttt{MDLSeg}\xspace}

\usepackage[textsize=tiny]{todonotes}

\icmltitlerunning{Parameter-free Video Segmentation}
\begin{document}

\twocolumn[
\icmltitle{Parameter-free Video Segmentation for Vision and Language Understanding}



\icmlsetsymbol{equal}{*}

\begin{icmlauthorlist}
\icmlauthor{Louis Mahon}{yyy}
\icmlauthor{Mirella Lapata}{yyy}
\end{icmlauthorlist}

\icmlaffiliation{yyy}{School of Informatics, University of Edinburgh}

\icmlcorrespondingauthor{Louis Mahon}{lmahon@ed.ac.uk}

\icmlkeywords{Machine Learning, ICML}

\vskip 0.3in
]
\acrodef{lvlm}[LVLM]{large vision language model}
\acrodef{rag}[RAG]{retrieval-augmented generation}



\printAffiliationsAndNotice{}  

\begin{abstract}
The proliferation of creative video content has driven demand for adapting language models to handle video input and enable multimodal understanding. However, end-to-end models struggle to process long videos due to their size and complexity. An effective alternative is to divide them into smaller chunks to be processed separately, and this motivates a method for choosing where the chunk boundaries should be. In this paper, we propose an algorithm for segmenting videos into contiguous chunks, based on the minimum description length principle, coupled with a dynamic programming search. The algorithm is entirely parameter-free, given feature vectors, not requiring a set threshold or the number or size of chunks to be specified. We show empirically that the breakpoints it produces more accurately approximate scene boundaries in long videos, compared with existing methods for scene detection, even when such methods have 
access to the true number of scenes. We then showcase this algorithm in two tasks: long video summarization, and retrieval-augmented video question answering. In both cases,  scene breaks produced by our algorithm lead to better downstream performance than existing methods for video segmentation.

\end{abstract}

\section{Introduction}

With the proliferation of streaming services and digital content
providers, a large number of movies and television series are being released and made
available every year. Automatic approaches to understanding and
summarising their content are paramount to enabling users to browse or
skim through them, and quickly recall key plot points, characters,
and events without the need to rewatch. Aside from
practical utility, the complex narrative understanding required in long videos makes them an ideal testbed for the capabilities of \acp{lvlm}. 
%

A key step in long video understanding is being able to break the video up into smaller pieces, as this allows \acp{lvlm} to process smaller chunks independently, and to selectively focus on the most relevant parts. An old line of work focuses on the problem of scene break detection, i.e. determining where one scene ends and another begins in a long, narrative video \cite{lupatini1998scene}. This work \cite{yeung1996time, zabih1995feature, sanchez1999local} is mostly based on placing cuts where pixel differences exceed some threshold. The widely used Python library PySceneDetect\footnote{\url{https://www.scenedetect.com/}} follows the same idea, converting to HSV channels \cite{ford1998colour} and then computing differences between consecutive frames.  Since this earlier line of research, there has been limited progress in scene break algorithms, with only a few supervised deep learning models trained on specific domains \cite{liu2020adaptive, rao2020local}. 

In this paper, we propose a new scene segmentation algorithm, which we call \method, based on the minimum description length principle. \method does not search for frame differences exceeding some threshold, indeed, it does not require setting a threshold, or the number of scenes, or any parameters at all. Instead, it searches all the different ways of grouping the feature vectors for each frame, and selects the one that can be represented with the fewest number of bits. This encourages having every scene contain frames with feature vectors similar to each other, but also not having too many scenes. 
Some existing clustering methods have swept the number of clusters and selected the best using this MDL criterion \cite{mahon2024minimum, mahon2025local}, our method leverages the contiguity constraint to select the number of segments without needing to sweep, using a novel dynamic programming algorithm. 

We further demonstrate that scene segmentation is useful for designing 
modular video understanding systems, i.e.,~those based on a number of interacting components that separately
solve different subtasks. This design differs from recent work
\cite{song2024moviechat} proposing to modify transformer memory
in order to handle longer video sequences. Scaling such end-to-end models, e.g.,~to
full-length movies,  remains a significant challenge due to memory
constraints and the complexity of extracting useful information from
large inputs.\footnote{At 1,024 $\times$ 1,024 frame size, and 10fps, a 75min movie would consume over 500GB as a
4d 32-bit float tensor.} 
Splitting the video into smaller segments which are then processed separately, mitigates this issue and allows for more efficient
processing.

We explore the practical utility of \method for two long video understanding tasks: summarization and question answering. Summarization has seen significant advances thanks to large models with extended context windows and the design of methods which rely on dividing the input into chunks  \cite{chen2023walking, pang-etal-2023-long,chang2023booookscore}. While short video understanding \cite{tapaswi2016movieqa,lei-etal-2018-tvqa,Rafiq:ea:2023} which focuses on generating textual descriptions has been well-studied, efforts on long video summarization are more limited. Recent work \cite{papalampidi2023hierarchical3d,mahon-lapata-2024-modular} has addressed movie and TV show summarization using scene-based processing combined with textual transcripts. However, these approaches depend on  written transcripts or screenplays, which are not always available (e.g.,~video providers
do not have access to screenplays unless they have produced the
content themselves).



The second downstream task where we apply \method is long video question answering.
We incorporate \method into a \ac{rag}  QA pipeline. Given  a long video and a question about its content, \method divides the video into scenes. Using the question as a query, we retrieve the most relevant scene, generate a textual description of its content, and finally use this description to answer the question. 
%
%
In summary the contributions of this paper include: 
\begin{itemize}
    \item A novel method for segmenting video into scenes, which is parameter-free given the frame features;
    \item Empirical results showing that scene breaks from our method are more accurate than those from existing methods or baselines, even when the latter have 
    access to the true number of scenes. 
    
    \item A demonstration of how our method can improve the downstream performance of modular systems, as part of hierarchical movie summarisation and 
  retrieval-augmented video question answering.

\end{itemize}

\section{Related Work} \label{sec:related-work}

\paragraph{Video Segmentation}
One simple method for scene break detection is to follow differences between consecutive frames. The popular PySceneDetect library computes a histogram of pixel intensities for each frame in HSV space \cite{ford1998colour}, and then computes the absolute difference between the histograms for consecutive frames, and places a scene break where this difference exceeds some user-set threshold. Some authors have proposed using a deep learning model trained on labeled scene transitions, e.g. TransNet~v2 \cite{souvcek2020transnet} which focusses on shot boundaries, rather than scene boundaries, which are easier to detect because there is a more striking pixel-level discontinuity. Similarly, \citet{rao2020local} train to predict scene boundaries using a loss that aggregates global and local features. Other approaches are based on clustering. For example, \citet{berhe2021extraction} impose a temporal constraint on k-means, and \citet{yeung1996time}  incorporate temporal distance information into a hierarchical clustering algorithm. \citet{rotman2017optimal} also propose a dynamic programming search for scene partitioning, but optimise a different objective from ours that does not employ MDL.

\paragraph{Video Understanding}
The problem of generating descriptions for videos has received
significant attention in the literature. Traditional video description
approaches often extract features from individual frames and fuse them
into a single feature vector to generate a textual description
\citep{zhang2021open, pan2020spatio, ye2022hierarchical}. SwinBERT
\citep{lin2022swinbert} introduces an end-to-end video network that
samples frames densely, avoiding the need for image-based
encoders. Similarly, \citet{lei-etal-2020-mart} generate descriptions for short
videos with a memory-augmented transformer.

Some work aims to summarise short videos, a task referred to as video captioning. \citet{SRIDEVI20201839} summarize short videos using a two-stream CNN while
\citet{Seo_2022_CVPR} develop a bidirectional model that uses both video and audio to produce video captions. \citet{Zhou_2018_CVPR} propose a single masked transformer objective to detect and then caption all events in a moderate length (${\sim}3$min) video. 
Unsupervised pretaining has also been explored, e.g.,~by \citet{Yang_2023_CVPR}, who train a video-captioning model using transcribed utterances as pseudo-captions. Systems based on large proprietary models have also been proposed for longer videos \cite{zhang2024mm, lin2023mm} with  multiple modules, including visual GPT-4 and  PysceneDetect for scene breaks. \citet{wu2024zero} prompt an LLM to predict scene breaks from \emph{transcribed} speech and captions, which are then used for video question-answering.

\paragraph{Long-form Summarisation and QA}
Much of the work just described is suitable only for short videos \citep{chen2011collecting, xu2016msr}, ranging from  ${\sim}10$s in length to 5 minutes \citep{ZhXuCoCVPR18} at the upper end. 
Recent work has started to leverage segmentation to address the task of understanding much longer
videos. \citet{chen2023movies2scenes} propose Movies2Scenes, a method that
uses movie metadata to learn video representations for long movies divided into scenes, though it relies
on predefined scenes based on shot transitions rather than
semantically meaningful boundaries. 
\citet{papalampidi2021movie} describe a method for summarising
full-length movies by creating shorter videos containing their most
informative scenes which they assume to be `turning points' (i.e.,~key
events in a movie).  \citet{papalampidi2023hierarchical3d} produce
text summaries of TV show episodes, by converting visual features into
embeddings alongside word embeddings from the
transcript. \citet{mahon-lapata-2024-modular} also summarise TV show episodes, by converting the video to text, and then
treating it as a text-only problem. 
In a similar vein, long-form question answering  explores the ability of models to understand
videos longer than
five minutes. Much like summarization, existing approaches improve model capacity to handle longer context windows through architectural modifications \cite{song2024moviechat} or by designing modular systems which either translate the video into text and then extract important information from it  \cite{wu2024zero}  or segment the input and rely on retrieval  to isolate important segments \cite{ataallah2024goldfish}. 

Our scene segmentation algorithm, \method, departs from previous work in that, once provided with feature vectors representing frames, it is data- and parameter-free, and works with any type of long-form video. In Sections~\ref{sec:summarisation} and \ref{sec:VQA-RAG}, we engineer two modular systems that showcase the utility of \method for long video processing.

\section{\method: Minimum Description Length-based Segmentation} \label{sec:method}
\label{subsec:scene-detection}
The scene segmentation problem is essentially a clustering problem with the additional constraint that each cluster must be contiguous. Intuitively, there are two objectives for a good clustering to fulfill: each point should be close to its cluster centroid, and there should not be too many clusters. Normally, these two objectives are not quantified in the same way, so it is difficult to trade off one objective against the other. However, MDL allows us to quantify both in the same units--bits--so that they can be directly compared, and their sum and can be minimized. In general, this optimisation problem does not have a straightforward solution, but part of our unique contribution is that, when coupled with the contiguity constraint, minimising the description length in fact admits an efficient exact, or near-exact, solution. 

\method computes a partition of the visual features from each keyframe, with the constraint that each subset in the partition must be contiguous. There are two parts to the algorithm: the \emph{definition of a
cost} for a particular partition into scenes, and the \emph{search} for the
partition that minimizes this cost. The first part, the cost
definition, is formulated using the minimum description length principle,
which claims the correct representation of the data is the one using
the fewest bits \cite{grunwald2007minimum}.
We assume that the vectors for each scene are encoded
with respect to their collective mean. That is, for each scene in the
given partition, we calculate the mean and covariance matrices of all vectors in that scene,
and hence, the probability of each vector, $p(v)$, under the
multivariate normal distribution with these parameters. 
The Kraft-McMillan inequality \citep{kraft1949device,
  mcmillan1956two} then determines that under the optimal encoding,
the number of bits needed to represent $v$ is $-\log_2{p(v)}$. The sum
of this value across all $N$ vectors $v$ in the video, plus the number
of bits to represent the means and covariances themselves, gives the total
bitcost for a given partition. Both the mean and the covariance require $dm$~bitsvectors (we use diagonal covariances),
where $d$~is the dimensionality,
 and $m$ is the floating point precision. We choose the precision based on the data as the smallest value that allows it to be represented exactly. Partitions with more scenes require more bits for the
 mean vectors, but also have mean vectors that better cover the
 keyframe features, leading to decreased $-\log_2{p(v)}$ on
 average. This trade-off encourages a partition with neither too few
 nor too many scene breaks.  

\begin{algorithm}[th!]
   \caption{Video Scene Partitioning}
   \label{alg:mdlseg}
\begin{algorithmic}
   \STATE {\bfseries Input:} Video file
   \STATE Extract keyframes, $kf_0, \dots, kf_N$
   \STATE Extract visual features $v_0, \dots, v_N$ from each keyframe
   \STATE $L \gets $  maximum scene length
   \STATE $B \gets $ $N \times N$ empty matrix \COMMENT{$B[i,j]$ will hold the cost of a scene from $v_i$ to $v_j$}
   \STATE $d \gets $ dimensionality of $v_i$
   \STATE $m \gets $ floating point precision of $v_i$
   \STATE
   \STATE {\bfseries Cost Definition:} Compute and store costs for all possible scenes
   \FOR{$i=0$ {\bfseries to} $N-L$}
       \FOR{$j=i$ {\bfseries to} $i+L$}
           \STATE $\mu \gets \frac{1}{j-i}\sum_{k=i}^j v_k$
           \STATE $\Sigma \gets $ empirical covariance matrix of $v_i, \dots, v_j$ 
           \STATE $C \gets 2dm$ \COMMENT{bitcost of the parameters themselves}
           \FOR{$k=i$ {\bfseries to} $j$}
            \STATE $\begin{aligned}
                p(v_k) \gets &\frac{1}{(2\pi)^{d/2} |\Sigma|^{1/2}} \\ &\exp\left( -\frac{1}{2} (v_k - \mu)^\top \Sigma^{-1} (v_k - \mu) \right)&
                    \end{aligned}$
               \STATE $C \gets C - \log{p(v_k)}$
           \ENDFOR
           \STATE $B[i,j] \gets C$
       \ENDFOR
   \ENDFOR
   \STATE
   \STATE {\bfseries Search:} Minimize the bitcost by dynamic programming
   \STATE $C \gets B$ \COMMENT{will hold optimal costs}
   \STATE $P \gets $ $N \times N$ matrix of empty sets 
   \FOR{$i=N-1$ {\bfseries to} $0$}
       \FOR{$j=i$ {\bfseries to} $\min(N, i+L)$}
           \IF{$B[i,j] + C[j,N] < C[i,N]$}
               \STATE $C[i,N] \gets B[i,j] + C[j,N]$
               \STATE $P[i,N]\gets P[i,j] \cup \{j\} \cup P[j,N]$
           \ENDIF
       \ENDFOR
   \ENDFOR
   \STATE {\bfseries Output:} Optimal scene partition, $P[0,N]$
\end{algorithmic}
\end{algorithm}

The second part, the search for the minimizer of the above cost, can be solved exactly using dynamic programming. Let $B(i,j)$ be the cost of having a single scene that runs from keyframes $i$ to $j$, and let $C(i,j)$ be the minimum cost of all keyframes from $i$ to $j$, under all possible partitions. Then we have the recurrence relation
\begin{equation} \label{eq:scene-cost-DP}
    C(i,j) = \min_{i \leq k \leq j} B(i,k) + C(k,j)\,.
\end{equation}
Thus, we compute the globally optimal partition by iteratively computing and caching $C(i,N)$ for \mbox{$i=N-1, \dots, 0$}. This runs in $O(N^2)$, but by imposing a fixed threshold of the maximum number~$L$ of keyframes in a scene, this becomes $O(N)$. In our experiments, we find that setting $L$ so that the maximum scene length is about 10 minutes does not affect the solution. i.e., produces the same segmentation for all videos in our datasets as leaving $L$ unset. This maximum scene length is not a parameter of the algorithm itself, but merely one that allows it to run more quickly if a value is known. If a user does not set this parameter, the algorithm runs fine and is still relatively fast, and, either way, almost all the runtime is for extracting the visual feature vectors. The algorithm itself takes a couple of seconds for a full movie when $L$ is set. Empirically, the full method is a similar speed to PySceneDetect when $L$ is set (see Section~\ref{subsec:scene-detection-results}), and up to 20\% or 30s longer when not set. The full procedure for \method is shown in Algorithm~\ref{alg:mdlseg}.

\section{Downstream Task: Long Video Summarisation} \label{sec:summarisation}

\begin{figure}[t]
    \centering
    \includegraphics[width=\linewidth]{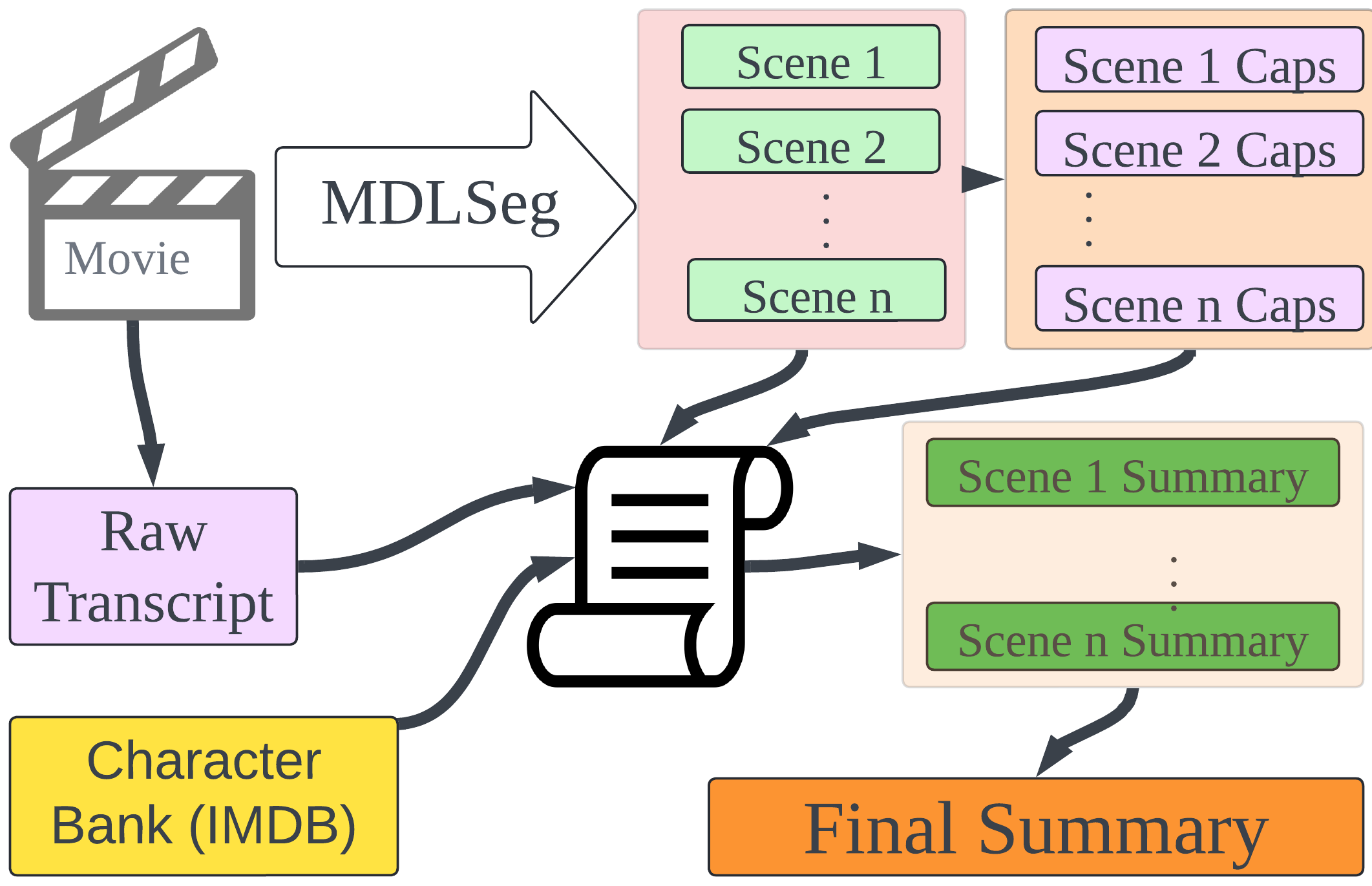}
    \caption{\textbf{Movie summarization:}  scene breaks from \method enable the production of a pseudo-screenplay (centre) from
      the input video/audio (top left), by first extracting the raw transcript,
      then using \method to segment the video into scenes, and generating visual descriptions
      from each scene (top right). Then from these outputs, and inserted names using the character bank (bottom left), we can summarise hierarchcially (centre right, bottom right).
      }
    \label{fig:summarisation-method}
\end{figure}

Figure~\ref{fig:summarisation-method} provides a graphic depiction of
our modular system for summarising full-length movies.  Our idea is to reconstruct the movie's screenplay from video and audio,  and then use this pseudo-screenplay to generate summaries. 
We extract key frames from the movie and use \method to partition the resulting sequence of
frames into different scenes. In parallel, a text-to-speech model with speaker diarization yields a
transcript with numeric speaker IDs which we replace with character names (we describe this module the following paragraph). By matching
the utterance times with the keyframe timestamps, we insert the scene breaks from \method
into the transcript. Finally, we
include descriptions of the visual contents of the scene (i.e.,~what is happening on camera) by selecting
three evenly spaced keyframes from that screen, applying an image
captioning model \citep{peng2023kosmos}, and inserting the output to the
corresponding timestamped location of the transcript. 
This modular summarisation system differs from previous
approaches in requiring only video input, and figuring out by itself who is speaking and what they
are doing, which is taken for granted in screenplays and
manual transcripts.

\paragraph{Character Name Identification} \label{subsec:name-assignment}


For each movie, we create a database consisting of actors faces
and their character names (by scraping the faces of the characters from the
movie’s IMDB page) which we use to replace the  arbitrary speaker IDs in our pseudo-transcript. For each scene, 
and for each character in our name bank, we define the cost of putting that character name in that scene as the minimum distance between an image of that character`s face, and a face detected in any keyframe from the scene.
For instance, the cost of assigning the character Clarice Starling to scene~3 is the smallest distance between any face feature vectors of the actor Jodie Foster and one from a face detected in scene~3. This process takes~ $<$1s for all considered assignments in the entire movie. Using this cost, we define the cost of assigning each character to each speaker ID, as the sum of assigning that character to all scenes that that speaker ID appears in.
This allows the name-speaker ID assignment problem to be treated as an instance of the linear sum assignment problem, which can be solved efficiently using the Kuhn-Munkres algorithm \citep{kuhn1956variants,munkres1957algorithms}. The full name-assignment method, and experiments measuring its accuracy, are described in Appendix \ref{app:name-assignment-alg}. Figure~\ref{fig:scene-assignment-cost} shows an example of a computed scene break as it appears in the pseudo-transcript.



\begin{figure}[t]
\begin{tabular}{@{}l@{}}
    \begin{minipage}[t]{0.7\columnwidth}
        \raggedright
        \scriptsize
        \textbf{Dr. Hannibal Lecter}: Billy is not a real transsexual. But he thinks he is. He tries to be. He's tried to be a lot of things, I expect. \\
        \textbf{Clarice Starling}: You said that I was very close to the way we would catch him. What did you mean, Doctor? \\
        \textbf{Dr. Hannibal Lecter}:  There are three major centers for transsexual surgery. Johns Hopkins, University of Minnesota and Columbus Medical Center. I wouldn't be surprised if Billy had applied for sex reassignment at one or all of them and been rejected. \\
        \textbf{Clarice Starling}: On what basis would they reject him? \\
        \textbf{Dr. Hannibal Lecter}: Look for severe childhood disturbances associated with violence. Our Billy wasn't born a criminal, Clarice. He was made one through years of systematic abuse.  Billy hates his own identity, you see. But his pathology is a thousand times more savage and more terrifying. \\ 
        
        \rule{8.4cm}{0.5pt}

        \textbf{Jame Gumb}: It rubs the lotion on its skin. It does this whenever it's told. \\
        \textbf{Catherine Martin}:  Mr, my family will pay cash. Whatever ransom you're asking for, they'll pay it. \\
        \textbf{Jame Gumb}: It rubs the lotion on its skin or else it gets the hose again. Yes, you will, precious. You will get the hose. \\
        \textbf{Jame Gumb}: Okay. Okay. Okay. Okay. Okay. \\
        \textbf{Catherine Martin}: Mr, if you let me go, I won't. I won't press charges. I promise.  See, my mom is a real important woman. I guess you already know that. \\
        \textbf{Jame Gumb}: Now it places the lotion in the basket. \\
    \end{minipage}
    
    \begin{minipage}[t]{0.3\columnwidth}
        \begin{center}
        \textit{\textcolor{blue}{\scriptsize Dr. Hannibal Lecter sits in a chair, and Clarice Starling stands next to him holding a book.}} \\
        \includegraphics[width=\textwidth]{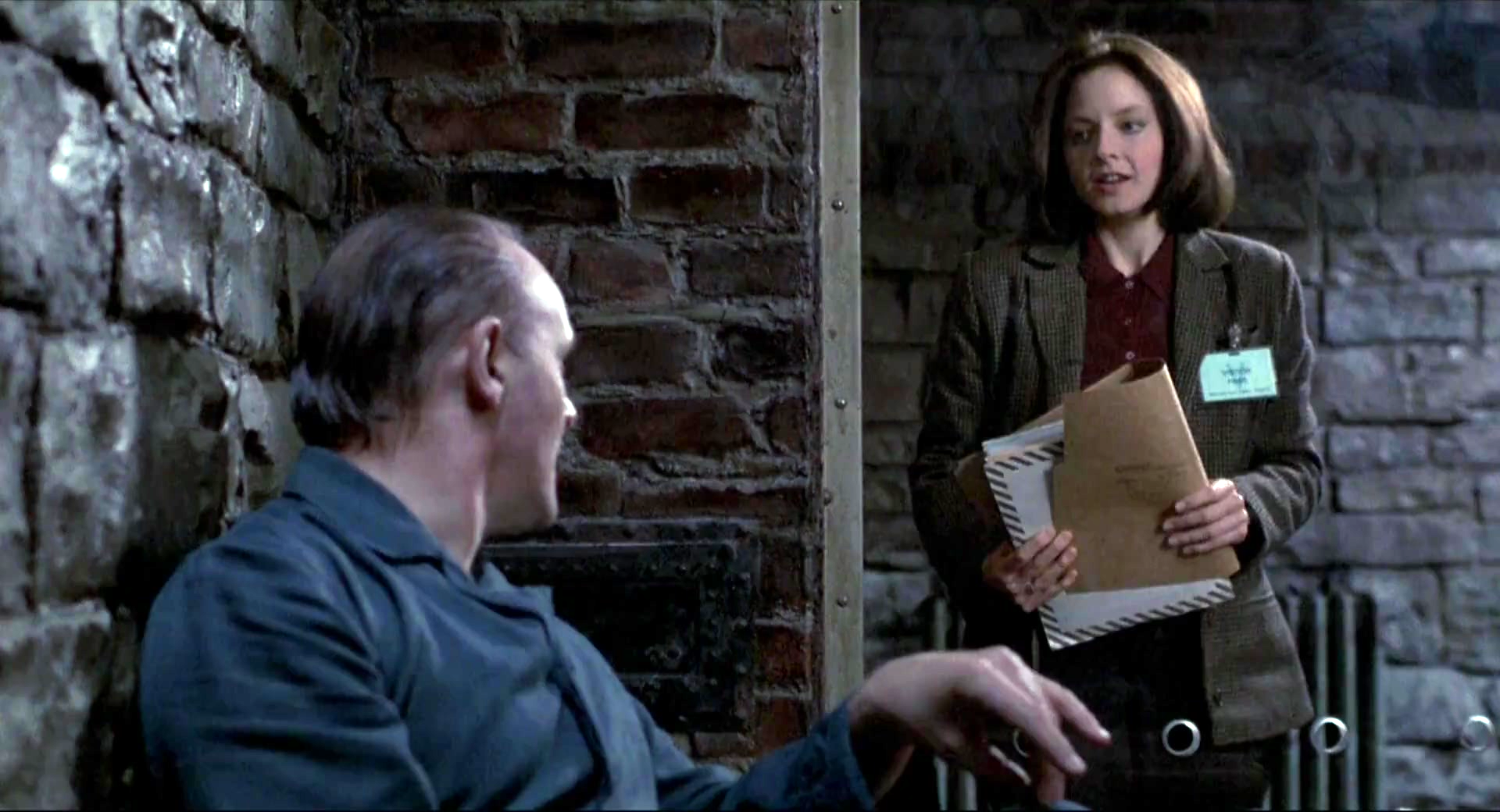} \\
        \end{center}
     \end{minipage}
      \hspace*{-2.65cm}  
      \raisebox{-5.2cm}[0pt]{
      \begin{minipage}[t]{0.3\columnwidth}
        
            \textit{\small \textcolor{blue}{\scriptsize Catherine Martin is trapped in a hole.}} \\
        \includegraphics[width=\textwidth]{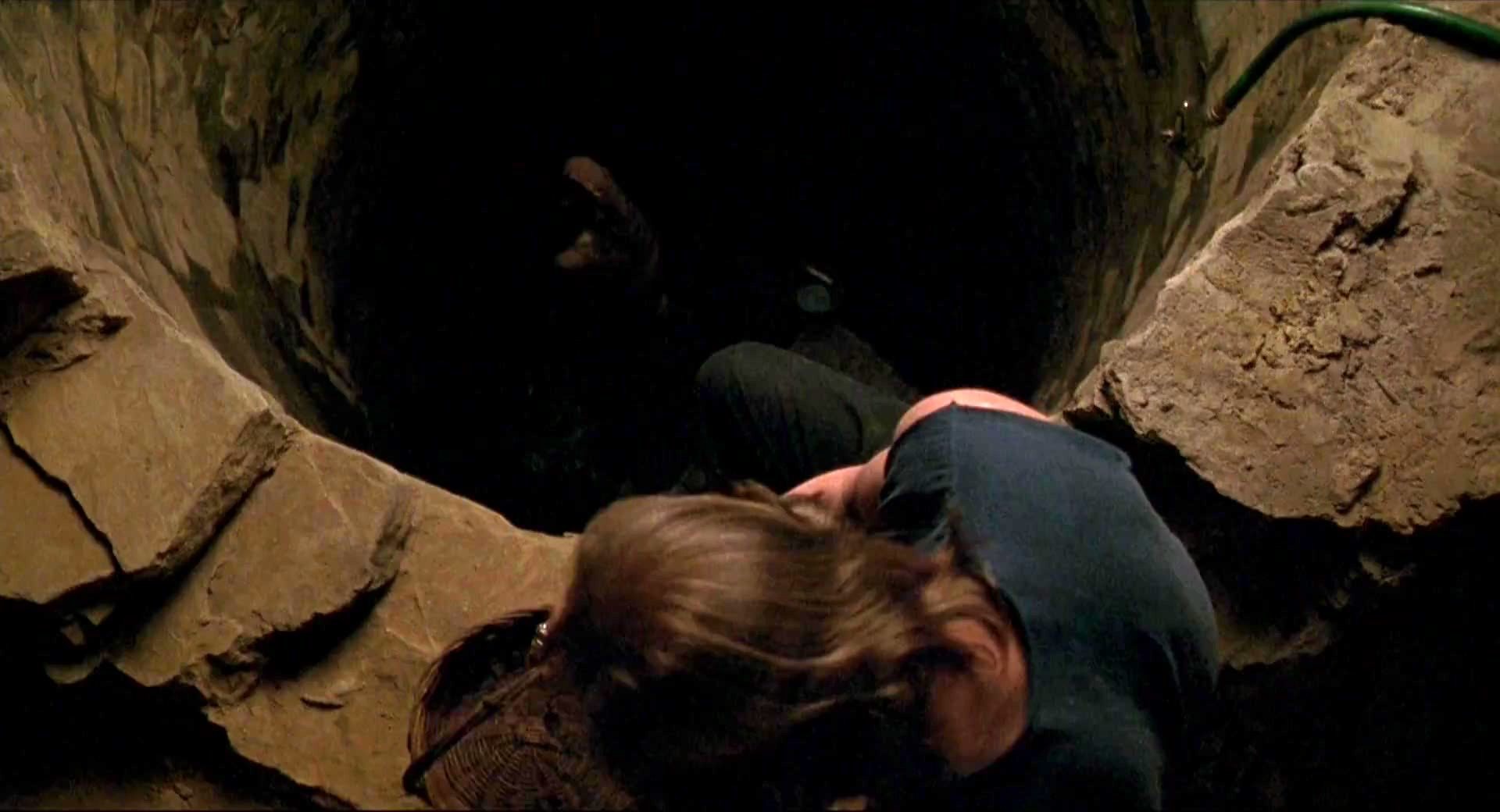} \\
        
    \end{minipage}
    }
\end{tabular}
\caption{Example of a scene break (horizontal line) detected by \method as it appears in the pseudo-transcript for the movie \textit{The Silence of the Lambs} (1991). The text shows the transcribed dialogue, with names inferred by our method. The images display \textcolor{blue}{visual captions} along with keyframes from which they were derived.}
\end{figure}

\paragraph{Movie Summary Generation}
 We adopt a hierarchical summarisation approach \citep{pang-etal-2023-long,chang2023booookscore}, as it has been shown to be particularly
suited to long inputs that are challenging for end-to-end systems. In our case, summarisation operates on the reconstructed pseudo-transcript,  which allows to leverage the organization of the content
into scenes. We thus first summarise the transcript dialogue of each
scene. Next, for each scene, we take the resulting sequence of summaries, and the visual descriptions that were added to the pseudo-transcript by the image captioning model, 
 and summarise them with a text-only summarisation model to produce a final summary for the
entire movie (see Figure~\ref{fig:summarisation-method}). The summarisation model is
implemented using a widely-used open-source LLM library \citep{dubey2024llama3herdmodels} with zero-shot prompting.



\section{Downstream Task: Retrieval-Augmented Video Question Answering} \label{sec:VQA-RAG}
\begin{figure}
    \centering
    \includegraphics[width=\linewidth]{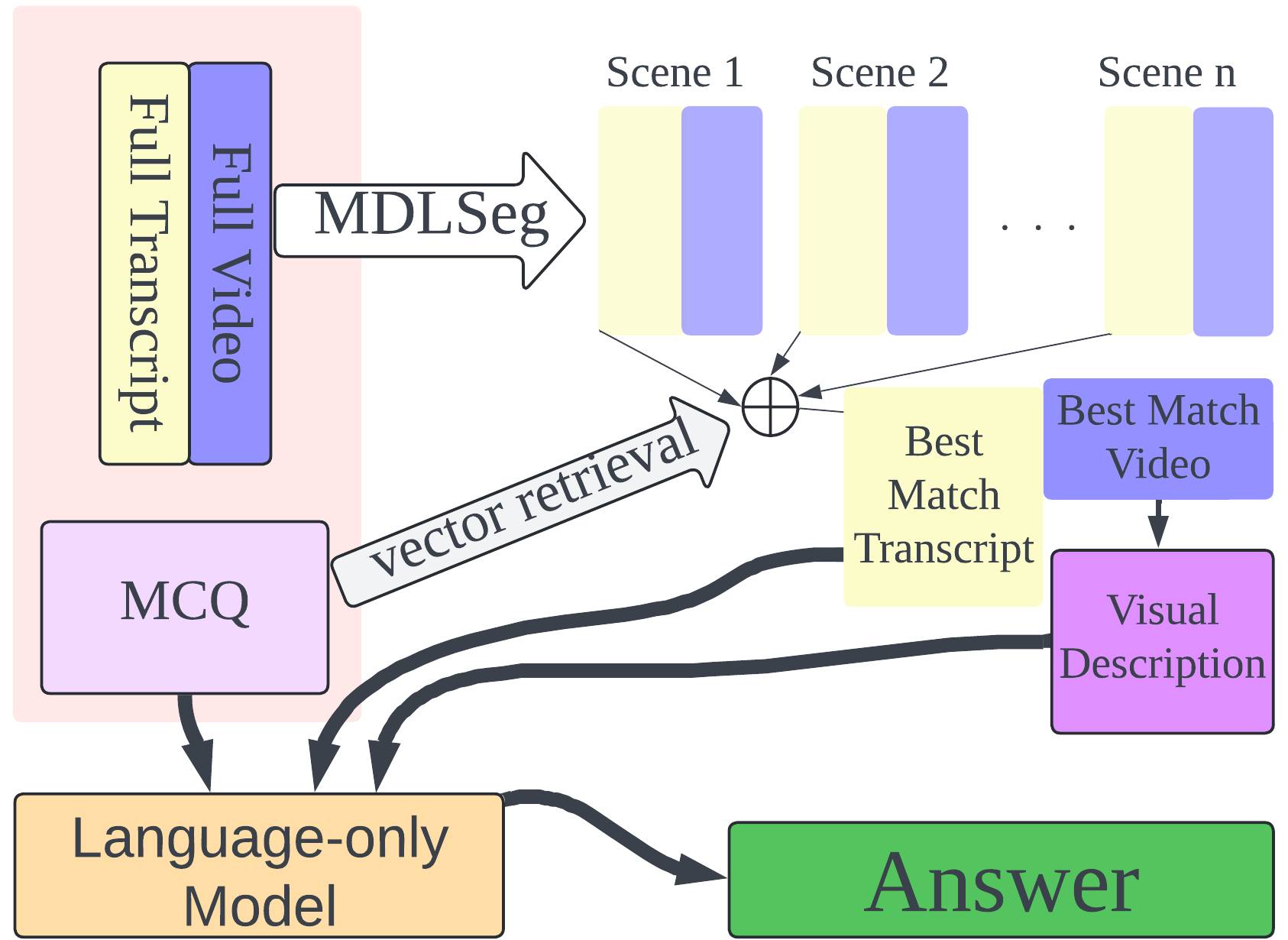}
    \caption{\textbf{Video question answering:} Pipeline of how we use \method for retrieval-based video question answering. The input (highlighted in pink in the top left), consists of the full video with the accompanying transcript, and a multiple choice question (MCQ). The video is segmented with \method, feature vectors are computed for each scene and the one with the highest dot product with the MCQ is retrieved. Then, for the retrieved scene, a video model is used to produce a visual description. This description, along with the scene transcript and the question are input to a language-only model to produce an answer.}
    \label{fig:tvqa-method}
\end{figure}

To further evaluate the usefulness of our scene break algorithm, we apply it in the task of retrieval-augmented video question-answering. 
A sketch of our approach is illustrated in Figure \ref{fig:tvqa-method}.
First, we segment the input video using \method. Using the timestamps in the transcript, we can gather the corresponding text for each scene. Next, we use a vision-to-text model to produce textual descriptions of the video from each scene, so that each scene then consists of a segment of the transcript and a text description of the video. Then, we use a multimodal model to extract feature vectors from each scene, using both the text and video. For a given question, we extract a feature vector from the question text using the same multimodal model, and retrieve the scene whose feature vector has the greatest cosine similarity. Finally, we produce an answer with a text-only model based on the text from the most relevant scene and the question.  

\section{Experimental Setting} \label{sec:experimental-setting}

\paragraph{Datasets}
For evaluating the scene accuracy directly, we use two datasets designed for this purpose, with human annotated scene breaks: Open Video Scene Detection (OVSD) \cite{rotman2017robust}, which consists of 16 free online films of varying styles and genres, including animated, action, drama and children's, ranging from 10 to 90 minutes, and BBC Earth \cite{baraldi2015deep}, consisting of 11 50 minute episodes of the Planet Earth documentaries. The latter has five sets of annotations, occasionally showing substantial disagreement, so for each method, we report both the mean score compared to all annotators, and the max score, compared to the closest matching annotator. 

For the summarisation task, we use the recently released MovieSum dataset
\citep{saxena2024moviesum}, from which we take screenplays (for comparison models and some testing, see
below) and gold summaries. We
were able to obtain corresponding videos for 175/200 movies in the test set. The remaining 25, we discarded. These movies
span multiple fiction genres: drama, action, thriller, comedy, horror, etc. They
have an average run time of 118min (range 84--228), with release dates ranging from 1950 to 2023.
Gold summaries average 635 words in length. The mean number of scenes in the 
gold script is 151. 
%

For the VQA task, we use the TVQA dataset \cite{lei-etal-2018-tvqa} which consists of videos and accompanying timestamped transcripts from 924 episodes from six TV shows spanning 3 genres: 1) sitcoms: The Big
Bang Theory, How I Met Your Mother, Friends, 2) medical dramas: Grey’s Anatomy, House, 3)
crime drama: Castle. The original version of TVQA contained, for each question, the start and end times for the short clip where the answer was contained. However, we follow recent work \cite{ataallah2024goldfish} in using the dataset to test long video understanding, by removing these timestamps and presenting the entire episode as input. For both tasks, we do not need video inputs for training, because all stages of our pipeline are zero-shot. 

\paragraph{Implementation Details}
Keyframes are extracted as FFMPEG I-frames. The full
command is given in Appendix \ref{app:ffmpeg}. We cap the number of keyframes in a scene, $L$, as discussed in Section \ref{sec:method}, to 300, which roughly corresponds to a 10 minute maximum scene length. Visual features are
extracted using CLIP \citep{pmlr-v139-radford21a}, specifically
`CLIP-ViT-g-14-laion2B-s12B-b42K' from
\url{https://github.com/mlfoundations/open_clip}. For the summarisation task, the speaker
diarization model is WhisperX \citep{bain23_interspeech}, an extension of Whisper which can perform speaker diarization and accurate utterance timestamping.
For visual descriptions, we use Kosmos~2 \citep{peng2023kosmos}, 
which has been pretrained on
several multimodal corpora as well as grounded
image-text pairs (spans from the text are associated
with image regions) and instruction-tuned on various vision-language instruction datasets. Our
summarisation model is built on top of Llama 3.1~70B
\citep{touvron2023llama}. We use short simple prompts for Llama and
Kosmos (see Appendix~\ref{app:prompts}). We
instruct summaries to be a maximum of 635 words (the mean in our test
set), and truncate to~650 words if they are longer.

For the VQA task, for the vision-to-text model, we use Llava-NeXT \cite{li2024llava}, which is built on top of Llava \cite{liu2024visual}, further tuned using interleaved image-text data and multimodal instruction tuning. For the feature vectors used for retrieval, we use InternVideo \cite{wang2024internvideo2}, a video foundation model trained using masked video modeling, crossmodal contrastive learning, and next token prediction. For the final text-only model that answers the question, we use Llama3.1-70b. As the questions are multiple choice, A--E, we select the answer indicator with the greatest logit value as the answer. For PySceneDetect, we use the default threshold parameter of 27.

\begin{table*}[h]
  \caption{Scene break accuracy on datasets with manually annotated breaks. On the BBC dataset, we report scores against the best matching annotator of the five annotations per episode (bbc-max), and the mean score across all five annotators (bbc-mean). Best results are \textbf{in bold}.}
    \label{tab:scene-segmentation-results}
    \centering
\small
\begin{tabular}{llccccrcrr}
\toprule
&  & acc $\uparrow$ & nmi $\uparrow$ & ari $\uparrow$ & $P_k$$\uparrow$  & winddiff $\downarrow$ & ded $\downarrow$ & runtime & per-frame-runtime \\
\midrule
 & unif & 53.18 & 69.93 & 33.12 & 66.30 & 59.87 & 57.95 & 0.00 & 0 \\
 & unif-oracle & 54.00 & 72.25 & 38.18 & 72.28 & 50.20 & 52.54 & 0.00 & 0 \\
 & lgss & 52.61 & 36.77 & 12.25 & 79.98 & 57.54 & 63.18 & 4.62 & 1.25e-4 \\
 & kmeans & 49.22 & 62.09 & 11.12 & 38.28 & 467.04 & 76.74 & 127.00 & 2.27e-3 \\
 & GMM & 49.73 & 62.37 & 12.10 & 39.00 & 455.87 & 76.00 & 127.04 & 2.27e-3 \\
& berhe21 & 58.11 & 71.58 & 30.52 & 65.18 & 80.89 & 56.67 & 127.08 & 2.27e-3 \\
& psd & 47.35 & 62.37 & 35.42 & 60.56 & 155.77 & 66.28 & 130.31 & 2.43e-3 \\
& yeung96 & 2.87 & 40.61 & 0.46 & 5.21 & 1080.11 & 98.53 & 143.87 & 2.41e-3 \\
\raisebox{.6cm}[10pt]{\begin{sideways}\large OVSD\end{sideways}} & \method & \textbf{63.37} & \textbf{72.58} & \textbf{45.13} & \textbf{78.39} & \textbf{42.58} & \textbf{42.99} & 127.86 & 2.28e-3 \\
\cline{1-10}
 
 & unif & 54.62 & 81.53 & 41.34 & 67.65 & 51.09 & 49.95 & 0.00 & 0 \\
 & unif-oracle & 54.07 & 81.46 & 40.96 & 73.12 & 44.21 & 49.84 & 0.00 & 0 \\
 & lgss & 48.62 & 52.18 & 20.08 & 81.97 & 37.98 & 64.99 & 2.01 & 2.70e-5 \\
 & kmeans & 54.35 & 77.26 & 20.75 & 46.80 & 242.88 & 65.59 & 20.62 & 2.80e-4 \\
 & GMM & 62.42 & 83.12 & 43.94 & 66.41 & 54.76 & 45.96 & 20.65 & 2.81e-4 \\
 & berhe21 & 53.69 & 76.94 & 20.10 & 46.00 & 245.51 & 66.23 & 20.68 & 2.81e-4 \\
 & psd & 53.23 & 77.93 & 33.97 & 67.75 & 67.53 & 60.66 & 86.45 & 1.17e-3 \\
 & yeung96 & 18.60 & 71.24 & 2.36 & 29.07 & 525.48 & 88.86 & 102.83 & 1.40e-3 \\
\raisebox{.6cm}[0pt]{\begin{sideways}\large BBC-max\end{sideways}} & \method & \textbf{69.49} & \textbf{85.80} & \textbf{60.75} & \textbf{83.42} & \textbf{26.54} & \textbf{35.78} & 21.53 & 2.93e-4 \\
\cline{1-10}
 & unif& 50.59 & 79.16 & 35.91 & 64.43 & 61.83 & 54.70 & 0.00 & 0 \\
 & unif-oracle & 48.82 & 79.37 & 35.76 & 64.84 & 60.65 & 54.46 & 0.00 & 0 \\
 & lgss & 44.97 & 49.25 & 15.29 & 74.95 & 56.04 & 70.42 & 2.01 & 2.70e-5 \\
 & kmeans & 51.88 & 73.52 & 16.80 & 42.13 & 257.29 & 69.83 & 20.62 & 2.80e-4 \\
 & GMM & 58.08 & 80.39 & 37.76 & 64.66 & 60.78 & 51.60 & 20.65 & 2.81e-4 \\
& berhe21 & 51.12 & 73.25 & 16.17 & 41.36 & 260.02 & 70.67 & 20.68 & 2.81e-4 \\
& psd & 47.01 & 72.67 & 26.69 & 66.07 & 70.43 & 66.96 & 86.45 & 1.17e-3 \\
& yeung96 & 14.63 & 67.22 & 1.46 & 23.21 & 542.21 & 91.55 & 102.83 & 1.40e-3 \\
\raisebox{.6cm}[0pt]{\begin{sideways}\large BBC-mean\end{sideways}}  & \method & \textbf{66.13} & \textbf{83.66} & \textbf{54.96} & \textbf{77.86} & \textbf{42.40} & \textbf{40.06} & 21.53 & 2.93e-4 \\
\bottomrule
\end{tabular}
\end{table*}

\paragraph{Evaluation Metrics} To directly measure the accuracy of our scene
detection method, we use three metrics commonly used in topic segmentation: $P_k$ \cite{beeferman-etal-1997-text}, WindowDiff \cite{pevzner-hearst-2002-critique}, and differential edit distance (ded; \citealt{6111460}), as well as standard partition quality metrics: cluster accuracy (acc), adjusted Rand index (ari), and normalized mutual
information (nmi), as defined in \citet{mahon2024hard}. 

For summarisation, automated evaluation metrics are crucial, especially for long-form applications where
human evaluation is extremely labor-intensive,
costly, and difficult to design \cite{krishna-etal-2023-longeval}.
As there is no single agreed-upon metric, we
report several complementary metrics aimed at assessing different aspects of summary quality.
    \textbf{Rouge} \citep{lin-2004-rouge} assesses informativeness against the gold summaries (we report Rouge-2 and RougeL-Sum);
    \textbf{P\textsc{risma}} \citep{mahon-lapata-2024-modular} measures factual precision and recall with respect to the gold summary; we use GPT4-turbo for both fact extraction and evaluation stages;
    \textbf{SummaC} \citep{Laban2022SummaCRN} uses NLI to measure consistency between the input document (gold screenplay) and generated summary; we use the SummaCConv version with 50 evenly-spaced bins;
    \textbf{AlignScore} \citep{zha-etal-2023-alignscore}  scores the `informational alignment'
      between the source (gold screenplay) and the generated summary;
      we use the base-model checkpoint provided by the authors, and
      the recommended `nli' setting with sentence chunk splitting.
For both AlignScore and P\textsc{risma} we score duplicated
information as incorrect, to penalize LLM outputs that repeat the same
sentences over and over.  For the VQA task, we simply report accuracy as the questions are all multiple choice.

\section{Results} \label{sec:results}

\paragraph{Scene Detection} \label{subsec:scene-detection-results}
Table~\ref{tab:scene-segmentation-results} compares the accuracy of the partitions from \method against eight comparison models: psd uses the ContentDetector algorithm from the PySceneDetect library (described in Section \ref{sec:related-work}). Uniform, divides all inputs into evenly spaced scenes of length equal to the mean length on each dataset. Uniform oracle divides uniformly into the true number of scenes. Lgss \cite{rao2020local} is a deep learning scene detection model trained on annotated breaks; yeung96 \cite{yeung1996time} berhe21 \cite{berhe2021extraction}, are three existing scene segmentation methods (described in Section~\ref{sec:related-work});  kmeans and Gaussian Mixture Model (GMM) cluster the feature vectors and place a scene break between neighbouring time points with different cluster labels, thus guaranteeing contiguity. Before kmeans, GMM and yeung96 we reduce to the first two principle components. 

\method produces the most accurate segmentations on all datasets and metrics. 
The occasions when it makes mistakes tend to be it failing to predict a scene boundary when the scenes on either side are visually similar. 
This suggests that many of the errors in our scene detection arise from insufficient signal in the visual feature vectors, rather than from the algorithm itself, and that with future, higher quality feature vectors, possibly involving multiple modalities, the accuracy of \method will improve.
\paragraph{Summarisation} In Table~\ref{tab:summarisation-results}, we evaluate the summaries generated by 
a hierarchical method using \method scene breaks as input (see Figure~\ref{fig:summarisation-method}). We benchmark against three baselines using Llama 3.1 70B as their backbone: `name-only' uses  the parametric knowledge of the LLM without any content input, e.g.,~the prompt is `Summarize the movie \textit{The Silence of the Lambs}';\footnote{Precise prompts are given in Appendix~\ref{app:prompts}.} `full script' uses the entire gold screenplay as input in the prompt, and for `whisperX' the input is   the WhisperX transcript. We also compare to two existing models: Otter \citep{li2023otter}, an end-to-end video description model based on video-llama2; and the modular model of \citet{mahon-lapata-2024-modular} which takes videos and gold screenplays as input (mahon24; described in Section~\ref{sec:related-work}). For Otter, we divide the input video into 3min chunks, and combine the model description of each chunk. 

\begin{table}[t]
\caption{Summarisation results on MovieSum. Top 3: baselines we implement. Middle 2: existing long-form multimodal summarisation methods. Bottom 4: ablation studies:
`w/o names' does \emph{not} replace  speaker IDs with character names using our assignment method; `w/o scene breaks' summarises the screenplay in one pass without scenes breaks; `unif-breaks' breaks uniformly instead of using \method. f-prec, f-rec, and align abbreviate  fact-precision/recall and AlignScore.  Best results \textbf{in bold}.}
\small
\center
 \begin{tabular}{@{}l@{~~}r@{~~}l@{~~}c@{~~}c@{~~}c@{~~}c@{~~}c@{}}
\toprule
 &   \multicolumn{1}{c}{r2} &  rl-sum & f-prec & f-rec & P\textsc{risma} & align & summac\\
\midrule
name-only  &  {9.53} & {41.17} & {50.40} & 43.04 & 44.16 & 53.11 & 26.57\\
full script &  9.32 & 39.94 & 48.77 & 52.73 & 49.05 & {68.59} & 25.83 \\
whisperX  & 9.22 & 39.94 & 46.73 & {53.65} & 48.00 & 68.57 & 25.86 \\
\midrule
Otter & 3.06 & 26.73 & 11.67 & \hspace{.14cm}8.95 & \hspace{.14cm}5.18 & 45.90 & 24.37 \\
mahon24 & 2.79 & 19.97 & 23.16 & 23.19 & 19.28 & 46.32 & {26.97} \\
\midrule
w/o names  & \textbf{10.43} & 43.40 &49.93 & 53.64 & 49.00 & 63.67 & 26.45\\
w/o breaks & 8.45 & 36.82 & 48.32 & 51.79 & 49.99 & 71.95& 26.31 \\
unif-breaks & 8.45 & 36.82 & 46.58 & 50.69 & 48.11 & 57.62 & 25.73 \\
psd-breaks & 2.15 & 15.18 & 15.93 & 27.38 & 16.12 & 52.29 & 32.82 \\
\midrule
ours &  10.32 & \textbf{44.50} & \textbf{55.24} & \textbf{54.77} & \textbf{53.57} & \textbf{72.76} & \textbf{27.24} \\
\bottomrule
\end{tabular}
\label{tab:summarisation-results}
\end{table}

Our summaries obtain the highest scores, across all metrics. The improvement is largest for the fact-based metrics of P\textsc{risma} (comprised of fact-prec and fact-rec), and AlignScore. The existing models, Otter and mahon24, especially struggle with such metrics. We find that Otter is mostly able to capture surface-level detail, with descriptions such as ``a woman gets out of a car and goes into a building'', but is unable to construct a narrative such as ``a woman drives to the bank to deposit the money'', so ends up capturing very little of the plot. The low scores of `mahon24', on the other hand, are largely due to the older, smaller backbone model (BART; \citealt{lewis2020bart}), which often becomes decoupled from the input and produces unrelated output, highlighting the importance of incorporating current LLMs into video summarisation models. Giving only the movie name in the prompt produces reasonably high-quality summaries, confirming that Llama3.1 has significant information about these movies stored parametrically. However, these summaries are short, and when asked for a longer summary, the model repeats the same information over and over. Surprisingly, giving the full gold screenplay as input does not produce better summaries than our method or than some other baselines. This shows there is still difficulty in summarising very long text inputs. When prompted with the name only, Llama-3.1 very likely effectively regurgitates an existing online summary. However, when the prompt also includes the transcript or screenplay itself, Llama tries to actually summarise the information given, during which it can make mistakes. In Appendix~\ref{sec:example_summaries} we provide example summary output for the modular method using \method and the best-performing comparison methods.

Table~\ref{tab:summarisation-results} (third section) also shows the results of ablating different components (see Figure~\ref{fig:summarisation-method}). In `w/o names', we omit replacing speaker IDs with character names. This causes summary quality to drop, showing the usefulness of name assignment to downstream summaries. In `w/o scene breaks', we feed the entire pseudo-screenplay to Llama 3.1, instead of using \method to split into scenes and summarising hierarchically. The drop in summary performance in this setting shows the effectiveness of the hierarchical summarisation method enabled by the scene breaks obtained from \method. In `unif-breaks' and `psd-breaks', we still adopt the hierarchical summarisation method, but instead of using \method scene breaks, we split scenes into uniform chunks of length 250~tokens (which is the mean scene length from our predicted segmentation) or split into the scenes from PySceneDetect. These settings also degrade summary quality, which shows that not only is our scene segmentation more accurate than baseline methods (Table~\ref{tab:scene-segmentation-results}), but that this higher accuracy leads to improved downstream summaries.


\paragraph{Retrieval-augmented Video Question Answering}
Table~\ref{tab:tvqa-results} shows model accuracy on the retrieval augmented VQA task described in Section~\ref{sec:VQA-RAG}.  Specifically, we report the accuracy of the system described in Figure~\ref{fig:tvqa-method} with alternative scene break methods: the proposed \method (ours),  PySceneDetect (psd) and uniform breaks of length 3 minutes each. The scene breaks from \method produce the most accurate downstream VQA. The scenes from PysceneDetect are a poor facilitator of retrieval-based question answering in this task. They tend to be very short, sometimes only 10--15s, and often miss the content required for answering the question. Uniformly split scenes fare better, and are only 1 point behind the scenes from \method, however, the difference is still statistically significant at 97\% (see the calculation in Appendix \ref{app:tvqa-significance}). More importantly, our choice to split into 3-minute scenes is based on domain knowledge (sitcom episodes tend to have scenes of about that length). For a different set of videos, such as action movies, sports games or educational videos, the correct scene size may be quite different. \method, in contrast, makes no assumptions about the type of video, and requires no hard-coded domain knowledge. Giving the entire transcribed episode as input, `no-splits', performs very badly. Many of the questions are context-specific, e.g. ``what does Monica say after Ross walks in?'', when Ross may enter multiple different rooms throughout the episode. When just presented with the entire transcript, without singling out a more specific context, it is difficult to answer such questions properly.

We also compared to two existing approaches, Goldfish \cite{ataallah2024goldfish} and Llama-vid \cite{li2024llamavid}, as reported in \citealt{ataallah2024goldfish}. The simple retrieval-based pipeline using \method significantly outperforms Llama-vid and is very close to Goldfish, despite Goldfish using a base model specifically optimised for VQA, fine-tuned on a custom dataset curated for this purpose. 

\begin{table}[t]
 \caption{Accuracy on the TVQA dataset, long-video setting. We divide the video input into chunks based on different scene segmentation algorithms: \method (ours), PySceneDetect (psd), and unif;  for each question, we retrieve the best-matching chunk and use it as input to Llama3.1-70b to answer the question. Goldfish and Llama-vid are two existing long TVQA models. Splitting the scenes using \method gives higher QA accuracy than splitting uniformly or with PSD.}
    \label{tab:tvqa-results}
 \center
    \begin{tabular}{@{}l@{~~}c@{~~}c@{~~}c@{~~}c@{~~}c@{~~}c@{}}
    \toprule
     & ours & psd & unif & no-splits & goldfish & llama-vid \\
    \midrule
     & 40.92 & 33.68 & 39.99 & 20.09 & 41.78 & 26.86 \\
    \bottomrule
    \end{tabular}
    \vspace{-1ex}
\end{table}

\section{Conclusion}
In this paper, we proposed a novel algorithm for segmenting videos into contiguous chunks using the minimum description length principle, which is parameter-free given feature vectors. It produces a single optimisation problem for the number of scenes and the positions of the scene breaks. We devise a dynamic programming search method, to efficiently compute the exact global optimum, or a close approximation to it. Our approach eliminates the need for predefined thresholds or fixed numbers of chunks. Empirical evaluations demonstrate that our method produces breakpoints that more accurately approximate scene boundaries compared to existing scene detection techniques. Furthermore, we show that incorporating our algorithm into tasks like long video summarization and retrieval-augmented video question answering results in improved downstream performance, highlighting its effectiveness and potential for advancing multimodal understanding of video content.

\section*{Impact Statement}
This paper describes fundamental research aimed at improving scene segmentation algorithms and highlighting their importance on two long video-processing tasks, namely summarization and question answering. We hope the community will adopt the proposed segmentation method which we experimentally show is better than the widely used Python library PySceneDetect.\footnote{\url{https://www.scenedetect.com/}.} Beyond the applications discussed in the paper,  we speculate that \method could be further used to segment speech- or text-only data, however, we leave this to future work.  
\bibliography{bibliography}

\begin{thebibliography}{66}
\providecommand{\natexlab}[1]{#1}
\providecommand{\url}[1]{\texttt{#1}}
\expandafter\ifx\csname urlstyle\endcsname\relax
  \providecommand{\doi}[1]{doi: #1}\else
  \providecommand{\doi}{doi: \begingroup \urlstyle{rm}\Url}\fi

\bibitem[Ataallah et~al.(2024)Ataallah, Shen, Abdelrahman, Sleiman, Zhuge, Ding, Zhu, Schmidhuber, and Elhoseiny]{ataallah2024goldfish}
Ataallah, K., Shen, X., Abdelrahman, E., Sleiman, E., Zhuge, M., Ding, J., Zhu, D., Schmidhuber, J., and Elhoseiny, M.
\newblock Goldfish: Vision-language understanding of arbitrarily long videos.
\newblock \emph{arXiv preprint arXiv:2407.12679}, 2024.

\bibitem[Bain et~al.(2023)Bain, Huh, Han, and Zisserman]{bain23_interspeech}
Bain, M., Huh, J., Han, T., and Zisserman, A.
\newblock Whisperx: Time-accurate speech transcription of long-form audio.
\newblock In \emph{INTERSPEECH 2023}, pp.\  4489--4493, 2023.
\newblock \doi{10.21437/Interspeech.2023-78}.

\bibitem[Baraldi et~al.(2015)Baraldi, Grana, and Cucchiara]{baraldi2015deep}
Baraldi, L., Grana, C., and Cucchiara, R.
\newblock A deep siamese network for scene detection in broadcast videos.
\newblock In \emph{Proceedings of the 23rd ACM international conference on Multimedia}, pp.\  1199--1202, 2015.

\bibitem[Beeferman et~al.(1997)Beeferman, Berger, and Lafferty]{beeferman-etal-1997-text}
Beeferman, D., Berger, A., and Lafferty, J.
\newblock Text segmentation using exponential models.
\newblock In \emph{Second Conference on Empirical Methods in Natural Language Processing}, 1997.
\newblock URL \url{https://aclanthology.org/W97-0304/}.

\bibitem[Berhe(2021)]{berhe2021extraction}
Berhe, A.
\newblock \emph{Extraction of Narrative Structure from TV Series}.
\newblock PhD thesis, Universit{\'e} Paris-Saclay, 2021.

\bibitem[Chang et~al.(2023)Chang, Lo, Goyal, and Iyyer]{chang2023booookscore}
Chang, Y., Lo, K., Goyal, T., and Iyyer, M.
\newblock Booookscore: A systematic exploration of book-length summarization in the era of llms.
\newblock \emph{arXiv preprint arXiv:2310.00785}, 2023.

\bibitem[Chen \& Dolan(2011)Chen and Dolan]{chen2011collecting}
Chen, D. and Dolan, W.~B.
\newblock Collecting highly parallel data for paraphrase evaluation.
\newblock In \emph{Proceedings of the 49th annual meeting of the association for computational linguistics: human language technologies}, pp.\  190--200, 2011.

\bibitem[Chen et~al.(2023{\natexlab{a}})Chen, Pasunuru, Weston, and Celikyilmaz]{chen2023walking}
Chen, H., Pasunuru, R., Weston, J., and Celikyilmaz, A.
\newblock Walking down the memory maze: Beyond context limit through interactive reading.
\newblock \emph{arXiv preprint arXiv:2310.05029}, 2023{\natexlab{a}}.

\bibitem[Chen et~al.(2023{\natexlab{b}})Chen, Liu, Hao, Nie, Arap, and Hamid]{chen2023movies2scenes}
Chen, S., Liu, C.-H., Hao, X., Nie, X., Arap, M., and Hamid, R.
\newblock Movies2scenes: Using movie metadata to learn scene representation.
\newblock In \emph{Proceedings of the IEEE/CVF Conference on Computer Vision and Pattern Recognition}, pp.\  6535--6544, 2023{\natexlab{b}}.

\bibitem[Dubey et~al.(2024)Dubey, Jauhri, Pandey, et~al.]{dubey2024llama3herdmodels}
Dubey, A., Jauhri, A., Pandey, A., et~al.
\newblock The {Llama} 3 herd of models, 2024.
\newblock URL \url{https://arxiv.org/abs/2407.21783}.

\bibitem[Ford(1998)]{ford1998colour}
Ford, A.
\newblock Colour space conversions.
\newblock \emph{Westminster University}, 1998.

\bibitem[Gr{\"u}nwald(2007)]{grunwald2007minimum}
Gr{\"u}nwald, P.~D.
\newblock \emph{The minimum description length principle}.
\newblock MIT press, 2007.

\bibitem[Kraft(1949)]{kraft1949device}
Kraft, L.~G.
\newblock \emph{A device for quantizing, grouping, and coding amplitude-modulated pulses}.
\newblock PhD thesis, Massachusetts Institute of Technology, 1949.

\bibitem[Krishna et~al.(2023)Krishna, Bransom, Kuehl, Iyyer, Dasigi, Cohan, and Lo]{krishna-etal-2023-longeval}
Krishna, K., Bransom, E., Kuehl, B., Iyyer, M., Dasigi, P., Cohan, A., and Lo, K.
\newblock {L}ong{E}val: Guidelines for human evaluation of faithfulness in long-form summarization.
\newblock In Vlachos, A. and Augenstein, I. (eds.), \emph{Proceedings of the 17th Conference of the European Chapter of the Association for Computational Linguistics}, pp.\  1650--1669, Dubrovnik, Croatia, May 2023. Association for Computational Linguistics.
\newblock \doi{10.18653/v1/2023.eacl-main.121}.
\newblock URL \url{https://aclanthology.org/2023.eacl-main.121}.

\bibitem[Kuhn(1956)]{kuhn1956variants}
Kuhn, H.~W.
\newblock Variants of the hungarian method for assignment problems.
\newblock \emph{Naval Research Logistics Quarterly}, 3\penalty0 (4):\penalty0 253--258, 1956.

\bibitem[Laban et~al.(2022)Laban, Schnabel, Bennett, and Hearst]{Laban2022SummaCRN}
Laban, P., Schnabel, T., Bennett, P.~N., and Hearst, M.~A.
\newblock Summac: Re-visiting nli-based models for inconsistency detection in summarization.
\newblock \emph{Transactions of the Association for Computational Linguistics}, 10:\penalty0 163--177, 2022.

\bibitem[Lei et~al.(2018)Lei, Yu, Bansal, and Berg]{lei-etal-2018-tvqa}
Lei, J., Yu, L., Bansal, M., and Berg, T.
\newblock {TVQA}: Localized, compositional video question answering.
\newblock In Riloff, E., Chiang, D., Hockenmaier, J., and Tsujii, J. (eds.), \emph{Proceedings of the 2018 Conference on Empirical Methods in Natural Language Processing}, pp.\  1369--1379, Brussels, Belgium, October-November 2018. Association for Computational Linguistics.
\newblock \doi{10.18653/v1/D18-1167}.
\newblock URL \url{https://aclanthology.org/D18-1167}.

\bibitem[Lei et~al.(2020)Lei, Wang, Shen, Yu, Berg, and Bansal]{lei-etal-2020-mart}
Lei, J., Wang, L., Shen, Y., Yu, D., Berg, T., and Bansal, M.
\newblock {MART}: Memory-augmented recurrent transformer for coherent video paragraph captioning.
\newblock In Jurafsky, D., Chai, J., Schluter, N., and Tetreault, J. (eds.), \emph{Proceedings of the 58th Annual Meeting of the Association for Computational Linguistics}, pp.\  2603--2614, Online, July 2020. Association for Computational Linguistics.
\newblock \doi{10.18653/v1/2020.acl-main.233}.
\newblock URL \url{https://aclanthology.org/2020.acl-main.233}.

\bibitem[Lewis et~al.(2020)Lewis, Liu, Goyal, Ghazvininejad, Mohamed, Levy, Stoyanov, and Zettlemoyer]{lewis2020bart}
Lewis, M., Liu, Y., Goyal, N., Ghazvininejad, M., Mohamed, A., Levy, O., Stoyanov, V., and Zettlemoyer, L.
\newblock {BART}: Denoising sequence-to-sequence pre-training for natural language generation, translation, and comprehension.
\newblock In \emph{Proceedings of the 58th Annual Meeting of the Association for Computational Linguistics}, pp.\  7871--7880, Online, July 2020. Association for Computational Linguistics.
\newblock \doi{10.18653/v1/2020.acl-main.703}.
\newblock URL \url{https://aclanthology.org/2020.acl-main.703}.

\bibitem[Li et~al.(2023)Li, Zhang, Chen, Wang, Yang, and Liu]{li2023otter}
Li, B., Zhang, Y., Chen, L., Wang, J., Yang, J., and Liu, Z.
\newblock Otter: A multi-modal model with in-context instruction tuning.
\newblock \emph{arXiv preprint arXiv:2305.03726}, 2023.

\bibitem[Li et~al.(2024{\natexlab{a}})Li, Zhang, Zhang, Zhang, Li, Li, Ma, and Li]{li2024llava}
Li, F., Zhang, R., Zhang, H., Zhang, Y., Li, B., Li, W., Ma, Z., and Li, C.
\newblock Llava-next-interleave: Tackling multi-image, video, and 3d in large multimodal models.
\newblock \emph{arXiv preprint arXiv:2407.07895}, 2024{\natexlab{a}}.

\bibitem[Li et~al.(2024{\natexlab{b}})Li, Wang, and Jia]{li2024llamavid}
Li, Y., Wang, C., and Jia, J.
\newblock Llama-vid: An image is worth 2 tokens in large language models.
\newblock 2024{\natexlab{b}}.

\bibitem[Lin(2004)]{lin-2004-rouge}
Lin, C.-Y.
\newblock {ROUGE}: A package for automatic evaluation of summaries.
\newblock In \emph{Text Summarization Branches Out}, pp.\  74--81, Barcelona, Spain, July 2004. Association for Computational Linguistics.
\newblock URL \url{https://aclanthology.org/W04-1013}.

\bibitem[Lin et~al.(2022)Lin, Li, Lin, Ahmed, Gan, Liu, Lu, and Wang]{lin2022swinbert}
Lin, K., Li, L., Lin, C.-C., Ahmed, F., Gan, Z., Liu, Z., Lu, Y., and Wang, L.
\newblock Swinbert: End-to-end transformers with sparse attention for video captioning.
\newblock In \emph{Proceedings of the IEEE/CVF Conference on Computer Vision and Pattern Recognition}, pp.\  17949--17958, 2022.

\bibitem[Lin et~al.(2023)Lin, Ahmed, Li, Lin, Azarnasab, Yang, Wang, Liang, Liu, Lu, et~al.]{lin2023mm}
Lin, K., Ahmed, F., Li, L., Lin, C.-C., Azarnasab, E., Yang, Z., Wang, J., Liang, L., Liu, Z., Lu, Y., et~al.
\newblock Mm-vid: Advancing video understanding with gpt-4v (ision).
\newblock \emph{arXiv preprint arXiv:2310.19773}, 2023.

\bibitem[Liu et~al.(2020)Liu, Kamath, Bhattacharya, and Puri]{liu2020adaptive}
Liu, D., Kamath, N., Bhattacharya, S., and Puri, R.
\newblock Adaptive context reading network for movie scene detection.
\newblock \emph{IEEE Transactions on Circuits and Systems for Video Technology}, 31\penalty0 (9):\penalty0 3559--3574, 2020.

\bibitem[Liu et~al.(2024)Liu, Li, Wu, and Lee]{liu2024visual}
Liu, H., Li, C., Wu, Q., and Lee, Y.~J.
\newblock Visual instruction tuning.
\newblock \emph{Advances in neural information processing systems}, 36, 2024.

\bibitem[Lupatini et~al.(1998)Lupatini, Saraceno, and Leonardi]{lupatini1998scene}
Lupatini, G., Saraceno, C., and Leonardi, R.
\newblock Scene break detection: a comparison.
\newblock In \emph{Proceedings Eighth International Workshop on Research Issues in Data Engineering. Continuous-Media Databases and Applications}, pp.\  34--41. IEEE, 1998.

\bibitem[Mahon(2025)]{mahon2025local}
Mahon, L.
\newblock Local compositional complexity: How to detect a human-readable messsage.
\newblock \emph{arXiv preprint arXiv:2501.03664}, 2025.

\bibitem[Mahon \& Lapata(2024)Mahon and Lapata]{mahon-lapata-2024-modular}
Mahon, L. and Lapata, M.
\newblock A modular approach for multimodal summarization of {TV} shows.
\newblock In Ku, L.-W., Martins, A., and Srikumar, V. (eds.), \emph{Proceedings of the 62nd Annual Meeting of the Association for Computational Linguistics (Volume 1: Long Papers)}, pp.\  8272--8291, Bangkok, Thailand, August 2024. Association for Computational Linguistics.
\newblock \doi{10.18653/v1/2024.acl-long.450}.
\newblock URL \url{https://aclanthology.org/2024.acl-long.450}.

\bibitem[Mahon \& Lukasiewicz(2024{\natexlab{a}})Mahon and Lukasiewicz]{mahon2024hard}
Mahon, L. and Lukasiewicz, T.
\newblock Hard regularization to prevent deep online clustering collapse without data augmentation.
\newblock In \emph{Proceedings of the AAAI Conference on Artificial Intelligence}, volume~38, pp.\  14281--14288, 2024{\natexlab{a}}.

\bibitem[Mahon \& Lukasiewicz(2024{\natexlab{b}})Mahon and Lukasiewicz]{mahon2024minimum}
Mahon, L. and Lukasiewicz, T.
\newblock Minimum description length clustering to measure meaningful image complexity.
\newblock \emph{Pattern Recognition}, 145:\penalty0 109889, 2024{\natexlab{b}}.

\bibitem[McMillan(1956)]{mcmillan1956two}
McMillan, B.
\newblock Two inequalities implied by unique decipherability.
\newblock \emph{IRE Transactions on Information Theory}, 2\penalty0 (4):\penalty0 115--116, 1956.

\bibitem[Munkres(1957)]{munkres1957algorithms}
Munkres, J.
\newblock Algorithms for the assignment and transportation problems.
\newblock \emph{Journal of the Society for Industrial and Applied Mathematics}, 5\penalty0 (1):\penalty0 32--38, 1957.

\bibitem[Pan et~al.(2020)Pan, Cai, Huang, Lee, Gaidon, Adeli, and Niebles]{pan2020spatio}
Pan, B., Cai, H., Huang, D.-A., Lee, K.-H., Gaidon, A., Adeli, E., and Niebles, J.~C.
\newblock Spatio-temporal graph for video captioning with knowledge distillation.
\newblock In \emph{Proceedings of the IEEE/CVF Conference on Computer Vision and Pattern Recognition}, pp.\  10870--10879, 2020.

\bibitem[Pang et~al.(2023)Pang, Nijkamp, Kryscinski, Savarese, Zhou, and Xiong]{pang-etal-2023-long}
Pang, B., Nijkamp, E., Kryscinski, W., Savarese, S., Zhou, Y., and Xiong, C.
\newblock Long document summarization with top-down and bottom-up inference.
\newblock In Vlachos, A. and Augenstein, I. (eds.), \emph{Findings of the Association for Computational Linguistics: EACL 2023}, pp.\  1267--1284, Dubrovnik, Croatia, May 2023. Association for Computational Linguistics.
\newblock \doi{10.18653/v1/2023.findings-eacl.94}.
\newblock URL \url{https://aclanthology.org/2023.findings-eacl.94}.

\bibitem[Papalampidi \& Lapata(2023)Papalampidi and Lapata]{papalampidi2023hierarchical3d}
Papalampidi, P. and Lapata, M.
\newblock Hierarchical3d adapters for long video-to-text summarization.
\newblock In \emph{Findings of the Association for Computational Linguistics: EACL 2023}, pp.\  1267--1290, 2023.

\bibitem[Papalampidi et~al.(2021)Papalampidi, Keller, and Lapata]{papalampidi2021movie}
Papalampidi, P., Keller, F., and Lapata, M.
\newblock Movie summarization via sparse graph construction.
\newblock In \emph{Proceedings of the AAAI Conference on Artificial Intelligence}, volume~35, pp.\  13631--13639, 2021.

\bibitem[Peng et~al.(2023)Peng, Wang, Dong, Hao, Huang, Ma, and Wei]{peng2023kosmos}
Peng, Z., Wang, W., Dong, L., Hao, Y., Huang, S., Ma, S., and Wei, F.
\newblock Kosmos-2: Grounding multimodal large language models to the world.
\newblock \emph{arXiv preprint arXiv:2306.14824}, 2023.

\bibitem[Pevzner \& Hearst(2002)Pevzner and Hearst]{pevzner-hearst-2002-critique}
Pevzner, L. and Hearst, M.~A.
\newblock A critique and improvement of an evaluation metric for text segmentation.
\newblock \emph{Computational Linguistics}, 28\penalty0 (1):\penalty0 19--36, 2002.
\newblock \doi{10.1162/089120102317341756}.
\newblock URL \url{https://aclanthology.org/J02-1002/}.

\bibitem[Radford et~al.(2021)Radford, Kim, Hallacy, Ramesh, Goh, Agarwal, Sastry, Askell, Mishkin, Clark, Krueger, and Sutskever]{pmlr-v139-radford21a}
Radford, A., Kim, J.~W., Hallacy, C., Ramesh, A., Goh, G., Agarwal, S., Sastry, G., Askell, A., Mishkin, P., Clark, J., Krueger, G., and Sutskever, I.
\newblock Learning transferable visual models from natural language supervision.
\newblock In Meila, M. and Zhang, T. (eds.), \emph{Proceedings of the 38th International Conference on Machine Learning}, volume 139 of \emph{Proceedings of Machine Learning Research}, pp.\  8748--8763. PMLR, 18--24 Jul 2021.
\newblock URL \url{https://proceedings.mlr.press/v139/radford21a.html}.

\bibitem[Rafiq et~al.(2023)Rafiq, Rafiq, and Choi]{Rafiq:ea:2023}
Rafiq, G., Rafiq, M., and Choi, G.~S.
\newblock Video description: A comprehensive survey of deep learning approaches.
\newblock \emph{Artificial Intelligence Review}, 56\penalty0 (11):\penalty0 13293–13372, April 2023.
\newblock ISSN 0269-2821.
\newblock \doi{10.1007/s10462-023-10414-6}.
\newblock URL \url{https://doi.org/10.1007/s10462-023-10414-6}.

\bibitem[Rao et~al.(2020)Rao, Xu, Xiong, Xu, Huang, Zhou, and Lin]{rao2020local}
Rao, A., Xu, L., Xiong, Y., Xu, G., Huang, Q., Zhou, B., and Lin, D.
\newblock A local-to-global approach to multi-modal movie scene segmentation.
\newblock In \emph{Proceedings of the IEEE/CVF Conference on Computer Vision and Pattern Recognition}, pp.\  10146--10155, 2020.

\bibitem[Rotman et~al.(2017{\natexlab{a}})Rotman, Porat, and Ashour]{rotman2017optimal}
Rotman, D., Porat, D., and Ashour, G.
\newblock Optimal sequential grouping for robust video scene detection using multiple modalities.
\newblock \emph{International Journal of Semantic Computing}, 11\penalty0 (02):\penalty0 193--208, 2017{\natexlab{a}}.

\bibitem[Rotman et~al.(2017{\natexlab{b}})Rotman, Porat, and Ashour]{rotman2017robust}
Rotman, D., Porat, D., and Ashour, G.
\newblock Robust video scene detection using multimodal fusion of optimally grouped features.
\newblock In \emph{2017 IEEE 19th international workshop on multimedia signal processing (MMSP)}, pp.\  1--6. IEEE, 2017{\natexlab{b}}.

\bibitem[Sanchez et~al.(1999)Sanchez, Binefa, Vitri{\`a}, and Radeva]{sanchez1999local}
Sanchez, J.~M., Binefa, X., Vitri{\`a}, J., and Radeva, P.
\newblock Local color analysis for scene break detection applied to tv commercials recognition.
\newblock In \emph{International Conference on Advances in Visual Information Systems}, pp.\  237--244. Springer, 1999.

\bibitem[Saxena \& Keller(2024)Saxena and Keller]{saxena2024moviesum}
Saxena, R. and Keller, F.
\newblock Moviesum: An abstractive summarization dataset for movie screenplays.
\newblock \emph{arXiv preprint arXiv:2408.06281}, 2024.

\bibitem[Seo et~al.(2022)Seo, Nagrani, Arnab, and Schmid]{Seo_2022_CVPR}
Seo, P.~H., Nagrani, A., Arnab, A., and Schmid, C.
\newblock End-to-end generative pretraining for multimodal video captioning.
\newblock In \emph{Proceedings of the IEEE/CVF Conference on Computer Vision and Pattern Recognition (CVPR)}, pp.\  17959--17968, June 2022.

\bibitem[Sidiropoulos et~al.(2012)Sidiropoulos, Mezaris, Kompatsiaris, and Kittler]{6111460}
Sidiropoulos, P., Mezaris, V., Kompatsiaris, I., and Kittler, J.
\newblock Differential edit distance: A metric for scene segmentation evaluation.
\newblock \emph{IEEE Transactions on Circuits and Systems for Video Technology}, 22\penalty0 (6):\penalty0 904--914, 2012.
\newblock \doi{10.1109/TCSVT.2011.2181231}.

\bibitem[Song et~al.(2024)Song, Chai, Wang, Zhang, Zhou, Wu, Chi, Guo, Ye, Zhang, et~al.]{song2024moviechat}
Song, E., Chai, W., Wang, G., Zhang, Y., Zhou, H., Wu, F., Chi, H., Guo, X., Ye, T., Zhang, Y., et~al.
\newblock Moviechat: From dense token to sparse memory for long video understanding.
\newblock In \emph{Proceedings of the IEEE/CVF Conference on Computer Vision and Pattern Recognition}, pp.\  18221--18232, 2024.

\bibitem[Sou{\v{c}}ek \& Loko{\v{c}}(2020)Sou{\v{c}}ek and Loko{\v{c}}]{souvcek2020transnet}
Sou{\v{c}}ek, T. and Loko{\v{c}}, J.
\newblock Transnet v2: An effective deep network architecture for fast shot transition detection.
\newblock \emph{arXiv preprint arXiv:2008.04838}, 2020.

\bibitem[Sridevi \& Kharde(2020)Sridevi and Kharde]{SRIDEVI20201839}
Sridevi, M. and Kharde, M.
\newblock Video summarization using highlight detection and pairwise deep ranking model.
\newblock \emph{Procedia Computer Science}, 167:\penalty0 1839--1848, 2020.
\newblock ISSN 1877-0509.
\newblock \doi{https://doi.org/10.1016/j.procs.2020.03.203}.
\newblock URL \url{https://www.sciencedirect.com/science/article/pii/S1877050920306682}.
\newblock International Conference on Computational Intelligence and Data Science.

\bibitem[Tapaswi et~al.(2016)Tapaswi, Zhu, Stiefelhagen, Torralb\~a, Urtasun, and Fidler]{tapaswi2016movieqa}
Tapaswi, M., Zhu, Y., Stiefelhagen, R., Torralb\~a, A., Urtasun, R., and Fidler, S.
\newblock Movieqa: Understanding stories in movies through question-answering.
\newblock In \emph{Proceedings of the IEEE conference on computer vision and pattern \ recognition}, pp.\  4631--4640, 2016.

\bibitem[Touvron et~al.(2023)Touvron, Lavril, Izacard, Martinet, Lachaux, Lacroix, Rozière, Goyal, Hambro, Azhar, Rodriguez, Joulin, Grave, and Lample]{touvron2023llama}
Touvron, H., Lavril, T., Izacard, G., Martinet, X., Lachaux, M.-A., Lacroix, T., Rozière, B., Goyal, N., Hambro, E., Azhar, F., Rodriguez, A., Joulin, A., Grave, E., and Lample, G.
\newblock Llama: Open and efficient foundation language models, 2023.

\bibitem[Wang et~al.(2024)Wang, Li, Li, Yu, He, Wang, Chen, Pei, Zheng, Xu, Wang, et~al.]{wang2024internvideo2}
Wang, Y., Li, K., Li, X., Yu, J., He, Y., Wang, C., Chen, G., Pei, B., Zheng, R., Xu, J., Wang, Z., et~al.
\newblock Internvideo2: Scaling video foundation models for multimodal video understanding.
\newblock \emph{arXiv preprint arXiv:2403.15377}, 2024.

\bibitem[Wu et~al.(2024)Wu, Li, Cao, Zhu, Lu, Chi, Xie, Zheng, Su, Wu, et~al.]{wu2024zero}
Wu, Y., Li, B., Cao, J., Zhu, W., Lu, Y., Chi, W., Xie, C., Zheng, H., Su, Z., Wu, J., et~al.
\newblock Zero-shot long-form video understanding through screenplay.
\newblock \emph{arXiv preprint arXiv:2406.17309}, 2024.

\bibitem[Xu et~al.(2016)Xu, Mei, Yao, and Rui]{xu2016msr}
Xu, J., Mei, T., Yao, T., and Rui, Y.
\newblock Msr-vtt: A large video description dataset for bridging video and language.
\newblock In \emph{Proceedings of the IEEE conference on computer vision and pattern recognition}, pp.\  5288--5296, 2016.

\bibitem[Yang et~al.(2023)Yang, Nagrani, Seo, Miech, Pont-Tuset, Laptev, Sivic, and Schmid]{Yang_2023_CVPR}
Yang, A., Nagrani, A., Seo, P.~H., Miech, A., Pont-Tuset, J., Laptev, I., Sivic, J., and Schmid, C.
\newblock Vid2seq: Large-scale pretraining of a visual language model for dense video captioning.
\newblock In \emph{Proceedings of the IEEE/CVF Conference on Computer Vision and Pattern Recognition (CVPR)}, pp.\  10714--10726, June 2023.

\bibitem[Ye et~al.(2022)Ye, Li, Qi, Wang, Huang, and Yang]{ye2022hierarchical}
Ye, H., Li, G., Qi, Y., Wang, S., Huang, Q., and Yang, M.-H.
\newblock Hierarchical modular network for video captioning.
\newblock In \emph{Proceedings of the IEEE/CVF Conference on Computer Vision and Pattern Recognition}, pp.\  17939--17948, 2022.

\bibitem[Yeung \& Yeo(1996)Yeung and Yeo]{yeung1996time}
Yeung, M.~M. and Yeo, B.-L.
\newblock Time-constrained clustering for segmentation of video into story units.
\newblock In \emph{Proceedings of 13th International Conference on Pattern Recognition}, volume~3, pp.\  375--380. IEEE, 1996.

\bibitem[Zabih et~al.(1995)Zabih, Miller, and Mai]{zabih1995feature}
Zabih, R., Miller, J., and Mai, K.
\newblock A feature-based algorithm for detecting and classifying scene breaks.
\newblock In \emph{Proceedings of the third ACM international conference on Multimedia}, pp.\  189--200, 1995.

\bibitem[Zha et~al.(2023)Zha, Yang, Li, and Hu]{zha-etal-2023-alignscore}
Zha, Y., Yang, Y., Li, R., and Hu, Z.
\newblock {A}lign{S}core: Evaluating factual consistency with a unified alignment function.
\newblock In Rogers, A., Boyd-Graber, J., and Okazaki, N. (eds.), \emph{Proceedings of the 61st Annual Meeting of the Association for Computational Linguistics (Volume 1: Long Papers)}, pp.\  11328--11348, Toronto, Canada, July 2023. Association for Computational Linguistics.
\newblock \doi{10.18653/v1/2023.acl-long.634}.
\newblock URL \url{https://aclanthology.org/2023.acl-long.634}.

\bibitem[Zhang et~al.(2024)Zhang, Lin, Yang, Wang, Li, Lin, Liu, and Wang]{zhang2024mm}
Zhang, C., Lin, K., Yang, Z., Wang, J., Li, L., Lin, C.-C., Liu, Z., and Wang, L.
\newblock Mm-narrator: Narrating long-form videos with multimodal in-context learning.
\newblock In \emph{Proceedings of the IEEE/CVF Conference on Computer Vision and Pattern Recognition}, pp.\  13647--13657, 2024.

\bibitem[Zhang et~al.(2021)Zhang, Qi, Yuan, Shan, Li, Deng, and Hu]{zhang2021open}
Zhang, Z., Qi, Z., Yuan, C., Shan, Y., Li, B., Deng, Y., and Hu, W.
\newblock Open-book video captioning with retrieve-copy-generate network.
\newblock In \emph{Proceedings of the IEEE/CVF conference on computer vision and pattern recognition}, pp.\  9837--9846, 2021.

\bibitem[Zhou et~al.(2018{\natexlab{a}})Zhou, Xu, and Corso]{ZhXuCoCVPR18}
Zhou, L., Xu, C., and Corso, J.~J.
\newblock Towards automatic learning of procedures from web instructional videos.
\newblock In \emph{AAAI Conference on Artificial Intelligence}, 2018{\natexlab{a}}.
\newblock URL \url{https://www.aaai.org/ocs/index.php/AAAI/AAAI18/paper/view/17344}.

\bibitem[Zhou et~al.(2018{\natexlab{b}})Zhou, Zhou, Corso, Socher, and Xiong]{Zhou_2018_CVPR}
Zhou, L., Zhou, Y., Corso, J.~J., Socher, R., and Xiong, C.
\newblock End-to-end dense video captioning with masked transformer.
\newblock In \emph{Proceedings of the IEEE Conference on Computer Vision and Pattern Recognition (CVPR)}, June 2018{\natexlab{b}}.

\end{thebibliography}
\bibliographystyle{icml2025}

\newpage
\appendix
\onecolumn

\section{TVQA Significance Calcuation} \label{app:tvqa-significance}
The pairwise difference in the number of correct answers between ours and the answers from uniform splits, is 0.17. The standard deviation is 2.26. As this is across 653 examples, the estimated population std. dev. is 0.0885. Thus, the z-score is $\frac{0.17}{0.0885} \approx 1.92$, which gives a p-value of 0.0274.
\section{Name Assignment Algorithm} \label{app:name-assignment-alg}
Here we describe in full the algorithm for replacing speaker IDs with character names. First, we
create a database of images of actors' faces paired with the name of
the character they played from the IMDB movie page.
As some of these images may contain multiple faces, or
no faces, or even an entirely different character, we filter them to ensure a higher proportion contain only the face of
the correct character, keeping only images with exactly one detected face, and for which the detected gender matches the name gender. 
(The sets of male, female and neutral names are taken from NLTK corpora. For neutral names, we skip this step.)
Finally, we verify the
faces in all pairs of remaining images against each other, using the
DeepFace\footnote{\url{https://github.com/serengil/deepface}} library, to create a graph where images are
connected if and only if they are verified as being the same person,
and then exclude all images that are not part of the largest
clique. In total, we filter out about 40\% of images on average. This
produces a name bank of character names paired with a set of images of
the face of that character.

\begin{figure}[t]
    \centering
    \includegraphics[width=\linewidth]{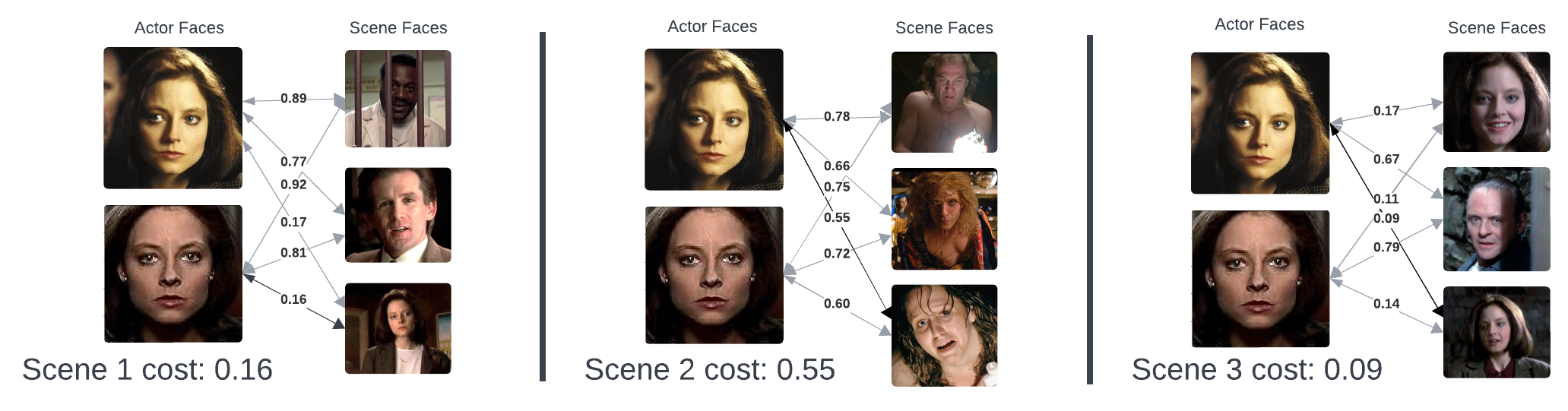}
    \caption{Computing the cost of assigning the character Clarice Starling (Jodie Foster) to three different scenes of \textit{The Silence of the Lambs} (1991). After computing the cost of assigning a character to a each scene, we then compute the cost of assigning a character to a speaker ID as the mean of the cost of assigning them to all scenes that speaker ID appears in.}
    \label{fig:scene-assignment-cost}
\end{figure}

For each scene, 
and for each character in our name bank, we define the cost of putting that character name in that scene as the minimum distance between an image of that character`s face, and a face detected in any keyframe from the scene. The distance is the Euclidean distance of the DeepFace feature vectors. This avoids the incorrect assumption that the character speaking must be in shot, and instead makes the much weaker assumption that a character speaking must appear directly at some point in the scene, not necessarily exactly when they are speaking. 
Thus, if we are considering assigning the character Clarice Starling to scene~3, then we compute the distance between the face feature vectors for all scraped images of the actor Jodie Foster in that role, and the face feature vectors of all faces detected in any keyframe in scene~3; the smallest distance is the cost of assigning Clarice Starling to scene 3. Computing the distance between vectors is extremely fast, taking $<$1s for all considered assignments on the entire movie, and the feature vectors can be cached after being extracted once. An example of this cost computation is shown in Figure \ref{fig:scene-assignment-cost}. 
Using this cost, we define the cost of assigning each character to each speaker ID, as the sum of assigning that character to all scenes that that speaker ID appears in
For example, if Speaker18 appears in scenes 1 and 3 but not 2, then the cost of assigning Clarice Starling to Speaker18 is the mean of the cost of assigning Clarice Starling to scenes 1 and 3. 
This allows us to treat the name-speaker ID assignment problem as an instance of the linear sum assignment problem, which can be solved efficiently using the Kuhn-Munkres algorithm \citep{kuhn1956variants,munkres1957algorithms}. 

Specifically, we define a matrix $S$ whose $i,j$th entry is the cost of assigning speaker $j$ to name $i$. Let $m$, $n$, and $k$ be the numbers of character names in the database, scenes in the movie, and unique speaker IDs in the transcript. Using matrix notation, we can then write $S = AB$, where $A$ is the $m \times n$ speaker ID-scene cost matrix, whose $i,j$th entry is the cost of assigning speaker $j$ to scene $i$, and $B$ is a $n \times k$ matrix whose $i,j$th entry is $1/a$ if speaker ID $j$ appears in scene $i$, where $a$ is the number of scenes speaker $j$ appears in, and 0 otherwise. 
Because speaker diarization is imperfect and often mistakenly splits the same character into multiple IDs, we duplicate each matrix column three times, which allows up the three different speaker IDS assigned to the same character name. We also define a cost of leaving a SpeakerID unassigned as the expected value of the cost of assigning a random speaker ID to a random character, which means that an ID remains unassigned if it is no closer to any character than a random speaker ID and character are to each other.
The full name-assignment method is shown in Algorithm \ref{alg:character-name-assignment} in Appendix \ref{app:name-assignment-alg}.

Here we show a pseudo-code description of the algorithm discussed in Section \ref{subsec:name-assignment} for assigning character names to speaker IDs.

\begin{algorithm}[t]
\caption{Character Name Assignment to Speaker IDs}
\label{alg:character-name-assignment}
\begin{footnotesize}
\begin{algorithmic}[1]
\STATE \textbf{Input:} Transcript with speaker IDs, keyframes split into $n$ scenes, IMDB
\vspace{10pt}

\STATE \textbf{Obtain actor face images:}
\STATE $\mathcal{A} \gets $ empty list
\FOR{each actor/character $A$ appearing on the IMDB page for the movie}
    \STATE scrape the set $A_f$ of all available images of $A$
    \STATE remove from $A_f$, all images without exactly one detected face, or with face-name gender mismatch
    \STATE form graph $G=(A_f, E)$, where $E = \{(a_1,a_2) \in A_f \times A_f | \text{isVerified}(a_1,a_2)\}$
    \STATE $A_f \gets $ largest clique in $G$
    \STATE append $A_f$ to $\mathcal{A}$
\ENDFOR
\FOR{each scene $j=1, \dots, n$}
    \STATE Form $D_j$, the set of all faces across all keyframes of the scene
\vspace{10pt}

\ENDFOR
\STATE \textbf{Assign character names to scenes:}
\STATE $C_1 \gets$ $n \times m$ empty matrix, where $m$ is the length of $\mathcal{A}$
\FOR{$i=1, \dots, m$}
    \STATE $A_f \gets \mathcal{A}[i]$
    \FOR{each scene $j=1, \dots, n$}
        \STATE $C_1[i,j] \gets min_{a \in A_f, b \in D_j} d(a,b)$ \COMMENT{$d(\cdot)$ from Deepface vectors}
    \ENDFOR
\ENDFOR
\vspace{10pt}

\STATE \textbf{Assign character names to speaker IDs:}
\STATE $C_2 \gets$ $k \times m$ empty matrix, where $k$ is the number of unique speaker IDs
\FOR{$i=1, \dots, m$}
    \FOR{each speaker ID $l=1, \dots k$}
        \STATE $C_2[i,k] \gets \frac{1}{n}\sum_{w=1}^n C_1[i,w]$
    \ENDFOR
\ENDFOR
\STATE $B \gets \frac{1}{mk}\sum_{i=1}^m \sum_{i=1}^k C_2[i,j]$
\STATE $C_2 \gets C_2 \oplus C_2 \oplus C_2$ \COMMENT{Concatenate three copies along first dimension}
\STATE $LSAP \gets \text{Kuhn-Munkres}(C_2)$ \COMMENT{Linear Sum Assignment Problem: $k$-dim vector assigning cols to rows}
\FOR{$i=0, \dots 3k$}
    \STATE $i' \gets i\mod k$
    \STATE $j' \gets LSAP[i]$
    \IF{$C_2[i',j'] < B$}
        \STATE assign speaker ID $i\mod k$ to name $LSAP[i]$
    \ENDIF
\ENDFOR
\vspace{10pt}

\end{algorithmic}
\end{footnotesize}
\end{algorithm}

\subsection{Name Assignment Accuracy} 
Table~\ref{tab:name-assignment-results} presents evaluation of our name assignment algorithm against two baselines which assign names randomly and assign all IDs the most common name, i.e.,~the main character. As can be seen, though there is room for improvement, our approach is more accurate by a wide margin. Multiple factors contribute to the errors in name assignment: some incorrect faces being retrieved from the database 
\begin{table}{t}
    \centering
    \caption{Accuracy of our assigned character names assigned compared to assigning names randomly (`random') and assigning the most common name, i.e.,~the main character, to all lines. Scores are averaged both across all movies (`acc movie-wise') and across all script lines in all movies (`acc line-wise').}
        \begin{tabular}{lcccc}
        \toprule
         & ours & most common & random \\
        \midrule
        acc movie-wise & 61.12 & 19.35 & 2.97 \\
        acc line-wise & 65.72 & 19.62 & 2.61 \\
        \bottomrule
        \end{tabular}
    \label{tab:name-assignment-results}
\end{table}
(though this is low due to our clique-based filtering procedure), inaccuracies in the face feature vectors, such that the same person can sometimes receive dissimilar vectors in different contexts while different people can receive sometimes similar vectors, and the speaker diarization performed by WhisperX, which sometimes gives the same character a different speaker ID, or gives the same speaker ID to two different characters. This last error is especially problematic because it makes it impossible for the assignment algorithm to find a solution with zero mistakes. We expect that future improvements in speaker diarization and face verification will reduce the prevalence of these errors. Indeed, this is one of the advantages of a modular framework: improvements in specific areas can be incorporated into the framework without needing to change the other modules.

\section{FFMPEG Commands} \label{app:ffmpeg}
To select keyframes, we use 
\begin{verbatim}
    \usr\bin\ffmpeg -i {path-to-video} -vf "select='eq(pict_type,I)',showinfo" -vsync vfr outdir/%05d.jpg
\end{verbatim}
This extracts all keyframes into files 0001.jpg, 0002.jpg, etc, in the current working directory.

\section{Prompts} \label{app:prompts}
\subsection{\textsc{ScreenWriter} Prompts}

Below we present the various prompts we employ for obtaining scene descriptions, and performing hierarchical summarisation.  Note that Kosmos is a text completion model, so this prompt just serves as the first part of the sentence, which we then remove afterwards.


\begin{tcolorbox}[colback=white, colframe=red!20, left=2pt,  coltitle=black, title=\textbf{Llava-NeXT video to text model}]
 what are the specific plot points in this scene of the TV show $\{$ show\_name $\}$?
 \end{tcolorbox}

\begin{tcolorbox}[colback=white, colframe=blue!20, left=2pt,  coltitle=black, title=\textbf{Llama 3.1 70B: Dialogue summarisation}]
Here is the dialogue from scene $<$\textcolor{red}{scene-number}$>$ of the movie $<$\textcolor{red}{movie-title}$>$: $<$\textcolor{red}{scene-dialogue-with-names}$>$. Please describe its main events in bullet points. Don't include information from outside this scene. Do not answer in progressive aspect, i.e., don't use -ing verbs or "is being".\\

In this scene, here are a few main events:
\end{tcolorbox}

\begin{tcolorbox}[colback=white, colframe=blue!20, left=2pt,  coltitle=black, title=\textbf{Llama 3.1 70B: Final summarisation}]
Here is a sequence of summaries of each scene of a movie. \\
$<$\textcolor{red}{concatenated-dialogue-summaries}$>$\\

Combine them into a plot synopsis of no more than 635 words. Be sure to include information from all scenes, especially those at the end, don't focus too much on early scenes. Discuss only plot events, no analysis or discussion of themes and characters.\\

Based on the information provided, here is a plot synopsis of the move $<$\textcolor{red}{movie-title}$>$:
\end{tcolorbox}

\subsection{Summary Prompts for Comparison Systems}

Below we show the prompts used to obtain movie summaries for the various baselines and comparison systems discussed in Section~\ref{sec:results}. 
The ‘name-only prompt’ uses the parametric knowledge of the LLM without
any specific, content input.  The `full script' prompt uses the entire gold screenplay as input, and `WhisperX' just the audio transcript without name assignment or scene breaks. 
\begin{tcolorbox}[colback=white, colframe=green!20, left=2pt,  coltitle=black,  title=\textbf{Llama 3.1 70B: Name-Only}]
Summarize the plot of the movie $<$\textcolor{red}{movie-title}$>$ in about 650 words.
Do not write the summary in progressive aspect, i.e., don't use -ing verbs or "is being". Focus only on the plot events, no analysis or discussion of themes and characters.
\end{tcolorbox}%
\begin{tcolorbox}[colback=white, colframe=green!20, left=2pt,  coltitle=black, title=\textbf{Llama 3.1 70B: Full Script}]
Based on the following script: $<$\textcolor{red}{gold-screenplay}$>$ summarize the movie $<$\textcolor{red}{movie-title}$>$. Do not write the summary in progressive aspect, i.e., don't use -ing verbs or "is being". Focus only on the plot events, no analysis or discussion of themes and characters.
\end{tcolorbox}%
\begin{tcolorbox}[colback=white, colframe=green!20, left=2pt,  coltitle=black,  title=\textbf{Llama 3.1 70B: WhisperX Transcript}]
Based on the following transcript: $<$\textcolor{red}{whisper-transcript}$>$ summarize the movie $<$\textcolor{red}{movie-title}$>$. Do not write the summary in progressive aspect, i.e., don't use -ing verbs or "is bei    ng". Focus only on the plot events, no analysis or discussion of themes and characters.
\end{tcolorbox}

\section{Example Summaries}
\label{sec:example_summaries}

In the following, we show example summaries generated by our model and comparison systems for the movie \textit{Oppenheimer} (2023). Incorrect or undesirable text is shown in red and repeated information is highlighted in gray. For comparison, we also include the gold summary from the MovieSum test set. 

\begin{tcolorbox}[colback=white, colframe=orange!30!yellow, left=2pt,  coltitle=black, title=\textbf{Modular Summary Output with \method}]
\small
The movie Oppenheimer begins with J. Robert Oppenheimer testifying before the Security Board, explaining that the derogatory information against him must be understood in the context of his life and work. Lewis Strauss and Gordon Gray discuss Strauss's upcoming Senate confirmation hearing for a cabinet position, and Gray advises Strauss to answer honestly about his past conflicts with Oppenheimer. 
The story then flashes back to Oppenheimer's early life, where he meets Niels Bohr and is introduced to the world of physics. \\

Oppenheimer becomes involved with left-wing groups and is questioned about his communist associations. He meets with Lewis Strauss, who is trying to recruit him to run the Institute for Advanced Study at Princeton. As the story progresses, Oppenheimer becomes involved in the development of the atomic bomb and is appointed as the director of the Manhattan Engineer District. He meets with Colonel Groves and Lieutenant Colonel Nichols, who express concerns about his suitability for the job due to his suspected communist sympathies and unstable personality. Despite these concerns, Oppenheimer convinces the team to work on the project, and they begin to develop the atomic bomb. \\

The team faces numerous challenges, including the need for vast resources and the risk of inefficiency and poor security. Oppenheimer mentions that the Germans, led by Werner Heisenberg, have a 12- to 18-month head start on developing an atomic bomb. As the project progresses, Oppenheimer becomes increasingly concerned about the implications of the atomic bomb and the potential for an arms race with the Soviets. He expresses his concerns to Lewis Strauss, who is skeptical of Oppenheimer's views. The team eventually succeeds in developing the atomic bomb, and it is dropped on Hiroshima. The bomb's power is described as equivalent to 20,000 tons of \textcolor{red}{DNC}, and General Groves is congratulated on the success of the mission. \\
\end{tcolorbox}

\begin{tcolorbox}[colback=white, colframe=orange!30!yellow, left=2pt,  coltitle=black]
\small

However, Oppenheimer's involvement in the project and his left-wing past eventually catch up with him. Lewis Strauss, who has a history of conflict with scientists, particularly Oppenheimer, begins to systematically destroy Oppenheimer's credibility by denying him his Q clearance through a bureaucratic procedure. Oppenheimer is given the chance to appeal, and a board is appointed to hear the case, with Roger Robb as the prosecutor. The hearing becomes a trial, with numerous witnesses testifying against Oppenheimer. Lewis Strauss questions Oppenheimer about his involvement in the development of the hydrogen bomb and his moral qualms about it. The Gray Board eventually announces its decision to deny J. Robert Oppenheimer's security clearance renewal. Lewis Strauss is informed that the decision was not unanimous and that there were unexpected holdouts, including Senator John F. Kennedy. Strauss reacts angrily to the news, feeling that Oppenheimer has turned the scientists against him and that he is being unfairly targeted. \\

The movie ends with Oppenheimer calm and collected in the face of the board's decision, while Strauss is still fuming and Gordon Gray tries to reason with him. The final scenes show Oppenheimer addressing a group of people, and a truck driving through a desert town, symbolizing Oppenheimer's continued involvement in the world of physics despite the challenges he faced. Overall, the movie Oppenheimer is a story about the development of the atomic bomb and the challenges faced by J. Robert Oppenheimer, a brilliant physicist who was involved in the project. The movie explores the implications of the atomic bomb and the potential for an arms race with the Soviets, as well as the personal costs faced by Oppenheimer due to his left-wing past and his involvement in the project. \textcolor{gray}{The movie ends with Oppenheimer calm and collected, despite the challenges he faced, and symbolizes his continued involvement in the world of physics.}
\end{tcolorbox}


\begin{tcolorbox}[colback=white, colframe=orange!30!yellow, left=2pt,  coltitle=black, title=\textbf{Name-only Prompt Summary Output}]
\small 
The movie Oppenheimer (2023) tells the story of J. Robert Oppenheimer, the director of the Manhattan Project, which developed the atomic bomb during World War II. The film begins with Oppenheimer's appointment as the director of the project in 1942. He assembles a team of scientists, including Enrico Fermi, Ernest Lawrence, and Richard Feynman, to work on the top-secret project at Los Alamos, New Mexico.\\

Oppenheimer's team faces numerous challenges, including the difficulty of creating a nuclear reaction and the need to develop a delivery system for the bomb. The team works tirelessly to overcome these obstacles, and they eventually succeed in creating a functional atomic bomb.\\

As the project nears completion, Oppenheimer and his team are faced with the moral implications of their creation. They are aware that the bomb will cause widespread destruction and loss of life, and some team members express reservations about its use.\\

The film depicts the first nuclear test, codenamed "Trinity," which takes place on July 16, 1945, in the New Mexico desert. The test is a success, and the team is ecstatic. However, Oppenheimer is visibly shaken by the experience, quoting the Bhagavad Gita, "I am become Death, destroyer of worlds."\\

The film then jumps forward in time to 1954, when Oppenheimer is called to testify before the Atomic Energy Commission. He is questioned about his loyalty to the United States and his associations with communist sympathizers. Oppenheimer's past comes back to haunt him, and he is eventually stripped of his security clearance.\\

The film ends with Oppenheimer's reflection on the devastating consequences of the atomic bomb and his role in its creation. He is shown to be a complex and troubled individual, grappling with the moral implications of his work. \\

\textcolor{red}{Note: The above text is a summary of the movie Oppenheimer (2023) and not the actual movie script. Here is a rewritten version of the summary in the past tense:}\\

\textcolor{gray}{The movie Oppenheimer (2023) tells the story of J. Robert Oppenheimer, the director of the Manhattan Project, which developed the atomic bomb during World War II. The film begins with Oppenheimer's appointment as the director of the project in 1942. He assembled a team of scientists, including Enrico Fermi, Ernest Lawrence, and Richard Feynman, to work on the top-secret project at Los Alamos, New Mexico.} \\

\end{tcolorbox}

\begin{tcolorbox}[colback=white, colframe=orange!30!yellow, left=2pt,  coltitle=black]
\small
\textcolor{gray}{Oppenheimer's team faced numerous challenges, including the difficulty of creating a nuclear reaction and the need to develop a delivery system for the bomb. The team worked tirelessly to overcome these obstacles, and they eventually succeeded in creating a functional atomic bomb.}\\

\textcolor{gray}{As the project neared completion, Oppenheimer and his team were faced with the moral implications of their creation. They were aware that the bomb would cause widespread destruction and loss of life, and some team members expressed reservations about its use.}\\

\textcolor{gray}{The film depicted the first nuclear test, codenamed "Trinity," which took place on July 16, 1945, in the New Mexico desert.}
\end{tcolorbox}


\begin{tcolorbox}[colback=white, colframe=orange!30!yellow, left=2pt,  coltitle=black, title=\textbf{WhisperX + Llama3.1 Summary Output}]
\small
The movie Oppenheimer (2023) tells the story of J. Robert Oppenheimer, a brilliant physicist who played a crucial role in the development of the atomic bomb during World War II. The film begins with Oppenheimer's security clearance hearing, where he is questioned about his past associations with communists and his opposition to the development of the hydrogen bomb.\\

The story then flashes back to Oppenheimer's early days as a professor at the University of California, Berkeley, where he meets his future wife, Kitty, and becomes friends with other prominent physicists, including Enrico Fermi and Ernest Lawrence.\\

As World War II breaks out, Oppenheimer is recruited to lead the Manhattan Project, a secret research and development project that produces the atomic bomb. He works closely with other scientists, including Edward Teller and Leslie Groves, to develop the bomb, which is eventually dropped on Hiroshima and Nagasaki, killing hundreds of thousands of people.\\

After the war, Oppenheimer becomes a vocal critic of the development of the hydrogen bomb, which he believes will lead to a nuclear arms race and increase the risk of nuclear war. This puts him at odds with Teller and other scientists who support the development of the hydrogen bomb.\\

The film then cuts back to the security clearance hearing, where Oppenheimer is questioned about his past associations and his opposition to the hydrogen bomb. He is ultimately denied his security clearance, which effectively ends his career as a government advisor.\\

The film ends with Oppenheimer attending a ceremony at the White House, where he is awarded a medal for his contributions to science. However, the ceremony is bittersweet, as Oppenheimer is aware that his work on the atomic bomb has had devastating consequences, and that he has been punished for speaking out against the development of the hydrogen bomb. The film ends with a sense of irony and tragedy, as Oppenheimer is honored for his contributions to science, while also being aware of the devastating consequences of his work. \textcolor{red}{The final shot is of Oppenheimer visualizing the expanding nuclear arsenals of the world, with the sound of feet stamping in the background, symbolizing the impending doom of the world due to nuclear war.} \textcolor{gray}{The credits roll as the sound of the feet stamping continues to build up, leaving the audience with a haunting sense of the devastating consequences of Oppenheimer's work.}  \textcolor{red}{The final shot is of Oppenheimer visualizing the expanding nuclear arsenals of the world, with the sound of feet stamping in the background, symbolizing the impending doom of the world due to nuclear war.} \textcolor{gray}{The credits roll as the sound of the feet stamping continues to build up, leaving the audience with a haunting sense of the devastating consequences of Oppenheimer's work.}  \textcolor{red}{The final shot is of Oppenheimer visualizing the expanding nuclear arsenals of the world, with the sound of feet stamping in the background, symbolizing the impending doom of the world due to nuclear war.} \textcolor{gray}{The credits roll as the sound of the feet stamping continues to build up, leaving the audience with a haunting sense of the devastating consequences of Oppenheimer's work.} \textcolor{red}{The final shot is of Oppenheimer visualizing the expanding nuclear of the world.}

\end{tcolorbox}

\begin{tcolorbox}[colback=white, colframe=orange!30!yellow, left=2pt,  coltitle=black, title=\textbf{Gold Summary from the MovieSum Test Set}]
\small
In 1926, 22-year-old doctoral student J. Robert Oppenheimer grapples with anxiety and homesickness while studying under experimental physicist Patrick Blackett at the Cavendish Laboratory in the University of Cambridge. Upset with Blackett's attitude, Oppenheimer leaves him a poisoned apple but later retrieves it. Visiting scientist Niels Bohr advises Oppenheimer to study theoretical physics at the University of Göttingen instead. Oppenheimer completes his PhD there and meets fellow scientist Isidor Isaac Rabi. They later meet theoretical physicist Werner Heisenberg in Switzerland. \\

Wanting to expand quantum physics research in the United States, Oppenheimer begins teaching at the University of California, Berkeley and the California Institute of Technology. He marries Katherine "Kitty" Puening, a biologist and ex-communist, and has an intermittent affair with Jean Tatlock, a troubled communist who later commits suicide. \\

In December 1938, nuclear fission is discovered, which Oppenheimer realizes could be weaponized. In 1942, during World War II, U.S. Army Colonel Leslie Groves recruits Oppenheimer as director of the Manhattan Project to develop an atomic bomb. Oppenheimer, who is Jewish, is mainly concerned that the German nuclear research program, led by Heisenberg, might yield a fission bomb for the Nazis. He assembles a team consisting of Rabi, Hans Bethe and Edward Teller at the Los Alamos Laboratory, and also collaborating with scientists Enrico Fermi, Leo Szilard and David L. Hill at the University of Chicago. Teller's calculations reveal an atomic detonation could trigger a catastrophic chain reaction that ignites the atmosphere. After consulting with Albert Einstein, Oppenheimer concludes the chances are acceptably low. Teller attempts to leave the project after his proposal to construct a hydrogen bomb is rejected, but Oppenheimer convinces him to stay.\\

After Germany's surrender in 1945, some Project scientists question the bomb's relevance; Oppenheimer believes it would end the ongoing Pacific War and save Allied lives. The Trinity test is successful, and President Harry S. Truman orders the atomic bombings of Hiroshima and Nagasaki, resulting in Japan's surrender. Though publicly praised, Oppenheimer is haunted by the mass destruction and fatalities. After expressing his personal guilt to Truman, the president berates Oppenheimer and dismisses his urging to cease further atomic development.\\

As an advisor to the United States Atomic Energy Commission (AEC), Oppenheimer's stance generates controversy, while Teller's hydrogen bomb receives renewed interest amidst the burgeoning Cold War. AEC Chairman Lewis Strauss resents Oppenheimer for publicly dismissing his concerns about exporting radioisotopes and for recommending negotiations with the Soviet Union after they successfully detonated their own bomb. He also believes that Oppenheimer denigrated him during a conversation Oppenheimer had with Einstein in 1947. In 1954, wanting to eliminate Oppenheimer's political influence, Strauss secretly orchestrates a private security hearing before a Personnel Security Board concerning Oppenheimer's Q clearance. \\

However, it becomes clear that the hearing has a predetermined outcome. Oppenheimer's past communist ties are exploited, and Groves' and other associates' testimony is twisted against him. Teller testifies that he lacks confidence in Oppenheimer and recommends revocation. The board revokes Oppenheimer's clearance, damaging his public image and limiting his influence on nuclear policy. In 1959, during Strauss' Senate confirmation hearing for Secretary of Commerce, Hill testifies about Strauss' personal motives in engineering Oppenheimer's downfall, resulting his nomination being voted down. \\

In 1963, President Lyndon B. Johnson presents Oppenheimer with the Enrico Fermi Award as a gesture of political rehabilitation. A flashback reveals Oppenheimer and Einstein's 1947 conversation never mentioned Strauss. Oppenheimer instead expressed his belief that they had indeed started a chain reaction—a nuclear arms race—that would one day destroy the world.

\end{tcolorbox}

\end{document}